%% file: TPAMI-VCM-v15-arXiv.tex
\documentclass[10pt,journal,compsoc]{IEEEtran}
\usepackage{graphicx}
\usepackage{times}
\usepackage{epsfig}
\usepackage{graphicx}
\usepackage{amsmath}
\usepackage{amssymb}
\usepackage{overpic}
\usepackage{subfigure}
\usepackage{multirow}
\usepackage{color}
\usepackage{cite}
\usepackage{diagbox}
\usepackage{mathtools}
\usepackage{bbding}

\usepackage[colorlinks]{hyperref}
\usepackage[numbers,sort&compress]{natbib}
\newcommand{\tabincell}[2]{\begin{tabular}{@{}#1@{}}#2\end{tabular}}

\usepackage{color,soul}

\renewcommand{\raggedright}{\leftskip=0pt \rightskip=0pt plus 0cm}

\definecolor{darkcyan}{rgb}{0.0, 0.55, 0.55}
\newcommand{\ly}[1]{{\color{black} #1}}

\soulregister\cite7
\soulregister\ref7
\soulregister\pageref7

\begin{document}
	\title{	
		Video Coding for Machine: Compact Visual Representation Compression for Intelligent Collaborative Analytics
	}
	
	\author{
	Wenhan Yang$^\ast$, Haofeng Huang$^\ast$, Yueyu Hu$^\ast$, Ling-Yu Duan, Jiaying Liu
	\\
	
	\IEEEcompsocitemizethanks{
		\IEEEcompsocthanksitem The authors are with Peking University, Beijing 100871, China. E-mail: \{yangwenhan, huang6013, huyy, lingyu, liujiaying\}@pku.edu.cn.
	}
	\IEEEcompsocitemizethanks{
		\IEEEcompsocthanksitem $^\ast$ indicates equal contribution.
	}
	
}
	
	
	\markboth{}
	{Shell \MakeLowercase{\textit{et al.}}: Bare Demo of IEEEtran.cls for Computer Society Journals}
	
	\IEEEtitleabstractindextext{
		
		\begin{abstract}
			\raggedright 
			As an emerging research practice leveraging recent advanced AI techniques, \textit{e.g.} deep models based prediction and generation, \textbf{V}ideo \textbf{C}oding for \textbf{M}achines~(\textbf{VCM}) is committed to bridging to an extent separate research tracks of video/image compression and feature compression, and attempts to optimize compactness and efficiency jointly from a unified perspective of high accuracy machine vision and full fidelity human vision.
			%
			With the rapid advances of deep feature representation and visual data compression in mind,
			in this paper, we summarize VCM methodology and philosophy based on existing academia and industrial efforts.
			The development of VCM follows a general rate-distortion optimization, and the categorization of key modules or techniques is established 
			including features assisted coding, scalable coding, intermediate feature compression/optimization, and machine vision targeted codec, from broader perspectives of vision tasks, analytics resources, \textit{etc.}
			From previous works, it is demonstrated that, although existing works \ly{attempt to reveal the nature of scalable representation in bits when dealing with machine and human vision tasks},
			\ly{there remains a rare study in the generality of low bit rate representation, and accordingly how to support a variety of visual analytic tasks.}
			Therefore, we investigate a \ly{novel} \textit{visual information compression for the analytics taxonomy} problem to strengthen the capability of compact visual representations extracted from multiple tasks for visual analytics.
			A new perspective of task relationships versus compression is revisited.
			By keeping in mind the transferability among different machine vision tasks (\textit{e.g.} high-level semantic and mid-level geometry-related), we aim to support multiple tasks jointly at low bit rates.
			In particular, to narrow the dimensionality gap between neural network generated features extracted from pixels and a variety of machine vision features/labels (\textit{e.g.} scene class, segmentation labels), a codebook hyperprior is designed to compress the neural network-generated features.
			As demonstrated in our experiments, this new hyperprior model is expected to improve feature compression efficiency by estimating the signal entropy more accurately, 
			which enables further investigation of the granularity of abstracting compact features among different tasks.
		\end{abstract}
		
		\begin{IEEEkeywords}
			Video coding for machine, analytics taxonomy, compact visual representation, multiple tasks, codebook-hyperprior
	\end{IEEEkeywords}}
	
	\maketitle
	
	\IEEEpeerreviewmaketitle
	
	\IEEEraisesectionheading{
		\section{Introduction}}
	
	The fields of computer vision and image/video compression have gained great progress in recent decades.
	While the former aims at crossing the semantic gap and translating image/video pixel signals into high-level semantic understanding information, such as recognition~\cite{alexnet,vgg,resnet} or detection tasks~\cite{fastrcnn,yolo},
	the latter pursues compact representation of pixel signals to improve storage and transmission efficiency.
	Driven by different targets, these two domains are developing separately to a large extent, and are rarely put together in discussion in the earlier researches.

	In recent years, at the application end, the rapid emergence and prosperity of smart cities~\cite{gao2021retina} and the Internet of Things (IoT)~\cite{zhang2021iot} raise the challenges to the original development route of the two domains, but also bring opportunities for their joint exploration and optimization.
	In the face of big data and massive applications, the original paradigm based on pixel signal compression~\cite{wiegand2003tcsvt,sullivan2012tcsvt,ma2015spm} can no longer meet the requirements of efficient analysis.
	At the theoretical end, the fast development of deep generative and analytics models~\cite{CycleGAN2017,NIPS2014_5ca3e9b1,Karras_2020_CVPR,zamir2018taskonomy,lin2014microsoft} has broken the bi-directional connection barrier between pixel signals and features.
	At the same time, the continuous development of multi-task learning~\cite{Simon2021mtl,Yu2020bdd}, disentangled representation learning~\cite{locatello19a}, unsupervised/self-learning~\cite{xie2020propagate,wang2020enhancing} and other techniques have greatly expanded the depth and breadth of feature representation learning mechanisms.
	Researchers are increasingly concerned about and pursuing comprehensive performance of features in open scenarios.

	Therefore, in this context, coding compression and analysis techniques for machine vision, called \textbf{video coding for machine (VCM)}~\cite{duan2020tip,yang2021tmm}, have emerged to \textit{build an efficient joint compression and analytics framework upon the combination of deep generative models, analytics models and coding techniques.
	The framework is capable of obtaining accurate, compact and generalized feature representations learned from multiple tasks in an end-to-end manner to effectively support big data intelligent analytics for massive diverse applications}.
	In general, there are three paths that new VCM approaches in recent two years are developed along.
	The first branch stands on the basis of image/video coding and rebuilds the codecs towards machine vision, called \textit{Machine Vision Targeted Codec}~\cite{Suzuki2019icip,Yang2020acmmm,Hou2020cvpr}.
	For these methods, they offer analytics-friendly images/videos, which can achieve better analytics performance with low bit-rates.
	The second branch extends the route of dedicated feature compression to compressing deep intermediate features, including \textit{Intermediate Feature Compression}~\cite{chen2019acmmm,chen2020icip,Suzuki2020icip} and \textit{Optimization}~\cite{Alvar2019icip,Singh2020icip}.
	The former aims to reconstruct the pretrained deep features according to the feature fidelity constraint
	while the latter directly optimizes the deep intermediate feature and compression models jointly based on task-driven analytics losses.
	The third branch explores more collaborative operations between video and feature streams, optimizing the image/video coding efficiency towards human vision (categorized as \textit{Feature Assisted Coding}~\cite{Chen2019icassp,li2019adacompress}) or improve the performance of both coding performance and intelligent analytics towards both human and machine visions (categorized as \textit{Scalable Coding}~\cite{wang2019icip,Hoang2020cvprw,yan2020icip}).

	Besides the academic papers, efforts are made at the standardization end.
	In July 2019, MPEG Video Coding for Machines Ad-Hoc group began to develop the related video coding standards of ``highly-efficient video compression and representation for intelligent machine-vision or hybrid machine/human-vision applications''~\cite{wg22020white}. 
	%
	The proposals cover aspects of use cases, requirements, processing pipelines, plan for potential VCM standards, and the evaluation framework including machine-vision tasks, dataset, evaluation metrics, and anchor generation, \textit{etc}.
	A proportion of proposals targets to provide potential solutions of video coding for machine with compression and analytics gains.
	%
	JPEG AI\footnote{https://jpeg.org/jpegai/index.html} also called for proposals of learning-based coding standards in April 2021.
	A single-stream compressed representation is adopted to improve both subjective quality from the human perspective, and effective performance from the perspective of machines to support a wide range of applications, including visual surveillance, autonomous vehicles and devices.
	
	Some endeavors are put into adopting the idea of VCM into industrial practice and systems.
	Nvidia announced a new video conferencing platform for developers called Nvidia Maxine\footnote{https://developer.nvidia.com/maxine}, which is claimed to solve some of the most common problems in video calls. 
	Maxine uses Nvidia's GPUs to process calls in the cloud and enhance them with the help of artificial intelligence techniques.
	According to the related research paper~\cite{wang2021facevid2vid}, 
	a novel neural model is developed to synthesize a  video of a talking head from a key frame and the motion keypoints.
	%
	Facebook released their low bandwidth video-chat compression method~\cite{maxime2020arxiv} to reconstruct faces on the decoder side authentically with facial landmarks extracted at the encoder side.
	The method can run on iPhone 8 in real time and allow video calling at a few kilobits per second.
	Aliyun's Video Cloud standards and implementation team\footnote{https://segmentfault.com/a/1190000039858782} also launched an AI-aided conference video compression system that saves 40-65\% bitrate compared to the latest Versatile Video Coding~(VVC) standard for the same human-eye viewing quality in a lab testing scenario.
	In their work, the key frame is compressed by VVC while the Jacobian matrix is encoded for motion modeling.
	The advantages over VVC in terms of high definition and subjective quality are even more pronounced, providing more lifelike facial expressions at lower bit-rates.
	
	Although previously mentioned works improve the intelligent analytics performance via optimizing image/video streams, visual feature representations, or both of them jointly, there is still a blank of research in the design of the low-bit-rate representations for diverse or even unseen visual analytics. 
	Through reviewing existing VCM works, we present the necessity of modeling the novel information compression for analytics taxonomy.
	The formulations are provided to model the transferability among different machine vision tasks under the compression condition.
	%
	The exploration of the new problem naturally leads to a novel hyperprior model that estimates the entropy of the neural network~(NN)-generated features more accurately. 
	Under the framework, we investigate a series of problems related to multi-task feature representation compression and obtain abundant insights that inspire the community and future works.
	
	In summary, our contributions are as follows.
	\begin{itemize}
		\item 
		We review the state-of-the-art approaches of video coding for machine with a unified generalized rate-distortion (R-D) optimization formulation. 
		We illustrate all methods in five categories (features assisted coding, scalable coding, intermediate feature compression/optimization, and machine vision targeted codec) and study the impact of analytics resources, approach output, supported analytics tasks, \textit{etc.}, on the potential VCM related techniques.
		\item 
		The survey naturally reveals the research blank of supporting diverse visual analytics tasks with low-bit-rate representations.
		To fill in this blank, we propose to investigate the novel \textit{information compression for analytics taxonomy} problem with an adjusted formulation considering the transferability among different machine vision tasks under the compression condition.
		\item 
		The investigation aims to deduce a unified compressed feature for both high-level semantic-related tasks and mid-level geometry analytic tasks.
		A codebook-based hyperprior model is designed to narrow down the intrinsic dimensionality gap between the NN-generated features from pixels and machine vision features/labels.
		Thus, the new model can estimate the entropy of NN-generated features more accurately, which helps minimize the bit-rates but still efficiently support different machine vision tasks.
		\item 
		Under the proposed compression architecture, we further conduct a comprehensive discussion on the joint compression of visual data for a series of tasks.
		The empirical results demonstrate the feasibility that a series of tasks could be supported by a unified compressed representation.
		We also explore more potentials of the compressed representation, \textit{e.g.} supporting unseen tasks, and plateau bit-rate in different tasks.

	\end{itemize}
	
	The rest of our paper is organized as follows.
	Section~\ref{sec:survey} conducts a comprehensive literature survey for the development of VCM in recent years, 
	presenting how these works optimize the defined VCM R-D cost in different ways.
	Section~\ref{sec:analytics} shows our exploration in modeling and optimizing the multiple tasks' R-D costs.
	In Section~\ref{sec:exp}, experimental configurations and results are presented.
	In Section~\ref{sec:discussion}, we make tentative discussions on several open issues based on our proposed framework, which provides rich insights for future researches.
	The concluding remarks and potential future directions are given in Section~\ref{sec:conclusion}.
	
	\section{Progress Survey of Video Coding for Machine}
	\label{sec:survey}
	
	\subsection{Formulation of VCM}
	
	%
	%
	%
	The $L+1$ tasks are bundled with the labels $\mathbf{Y} = \left\{ Y_{0}, Y_{1}, ..., Y_{L} \right\}$,
	whose features are denoted by
	$\mathbf{F} = \left \{ F_{0}, F_{1}, ..., F_{L} \right\}$ extracted from the image $I$:
	\begin{align}
		\left\{ F_{i} \right\}_{i=0,1,...,L} & = E\left( I | \theta_E \right), \\
		\widehat{Y}_{i} & = A\left( F_{i} | \theta_A \right),
	\end{align}
	where $E\left( \cdot | \theta_E \right)$ is the feature extractor, and $A\left( \cdot | \theta_A \right)$ is the analytics model that maps the feature into the end task representation.
	$\widehat{Y}_{i}$ is the label prediction that targets to regress $Y_{i}$.
	%
	%
	We define $l_{i} \left(\widehat{Y}_i, Y_i\right)$ (short as $l_{i}$) as the performance measure of the task $i$ regarding the reconstructed feature $\widehat{F}_{i}$ and label $\widehat{Y}_{i}$ after lossy encoding and decoding.
	The \textbf{VCM problem} is formulated as\textit{ an objective function to maximize the multi-task performance while minimizing the bit-rate cost}:
	\begin{align}
		\label{eq:vcm1}
		\footnotesize
		\mathop {\text{argmax} }\limits_{\Theta} & \sum\limits_{0 \le i \le L} {{\omega _i}{l_i }}, \text{ \textit{s.t.} } \sum\limits_{0 \le i \le L} {{\omega _i}}  = 1, \nonumber  \\
		& \widetilde{B} \left( \left\{ R_{F_{i}} \right\}_{0 \le i \le L} \right)  \le {S_T}, 
	\end{align}
	where  ${\omega _i}$ is the weighting parameter to balance the importance among different tasks.
	${S_T}$ is the total bit-rate cost constraint.
	$R_{F_{i}}$ measures the bit-rate of the feature $F_{i}$.
	$\widetilde{B} \left( \left\{ R_{F_{i}} \right\}_{0 \le i \le L} \right)$ measures the minimal bit-rate after fully considering the feature dependency among $\left\{ F_{i} \right\}_{0 \le i \le L}$.
	Besides $E\left( \cdot | \theta_E \right)$ and $A\left( \cdot | \theta_A \right)$, a VCM system still includes:
	\begin{itemize}
		\item Compression model $C\left( \cdot | \theta_C \right)$ that maps the feature into bit-stream;
		\item Decompression model $D\left( \cdot | \theta_D \right)$ that projects the bit-stream back to the feature;
		\item Feature predictor $G\left( \cdot | \theta_G \right)$ that targets predicting the feature of a task based on the reconstructed features of other more abstracted tasks to squeeze out the feature redundancy among different tasks,
	\end{itemize}
	which help form up $\widetilde{B} \left( \cdot \right)$.
	$\Theta = \left\{ {\theta_E}, {\theta_C},{\theta_D}, {\theta_G}, {\theta_A} \right\}$ are the parameters of all modules.
	%
	%
	%
	
	%
	More details about the form of $\widetilde{B} \left( \cdot \right)$ and how to optimize Eq.~\eqref{eq:vcm1} in terms of $\Theta$ will be briefly discussed in the following subsections.
	
	%
	
	\begin{figure}[t]
		\centering
		\includegraphics[width=1\linewidth]{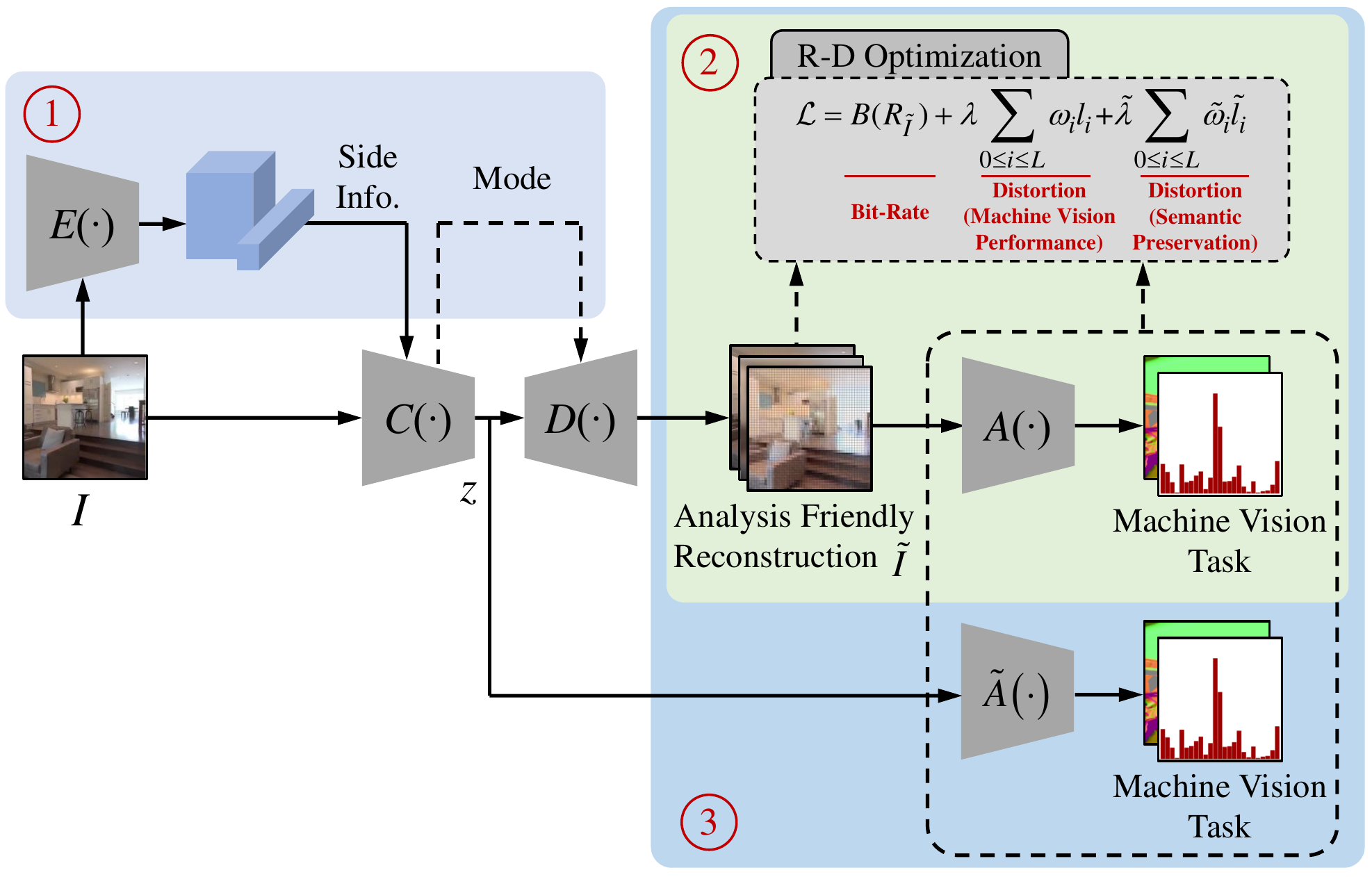}
		\caption{The framework of machine vision targeted codec, including: 1) side information guidance; 2) machine vision constraint; 3) semantic information preservation.}
		\label{fig:machine_targeted_codec}
	\end{figure}
	
	\begin{figure}[t]
		\centering
		\subfigure[Intermediate Feature Compression]{
			\includegraphics[width=1\linewidth]{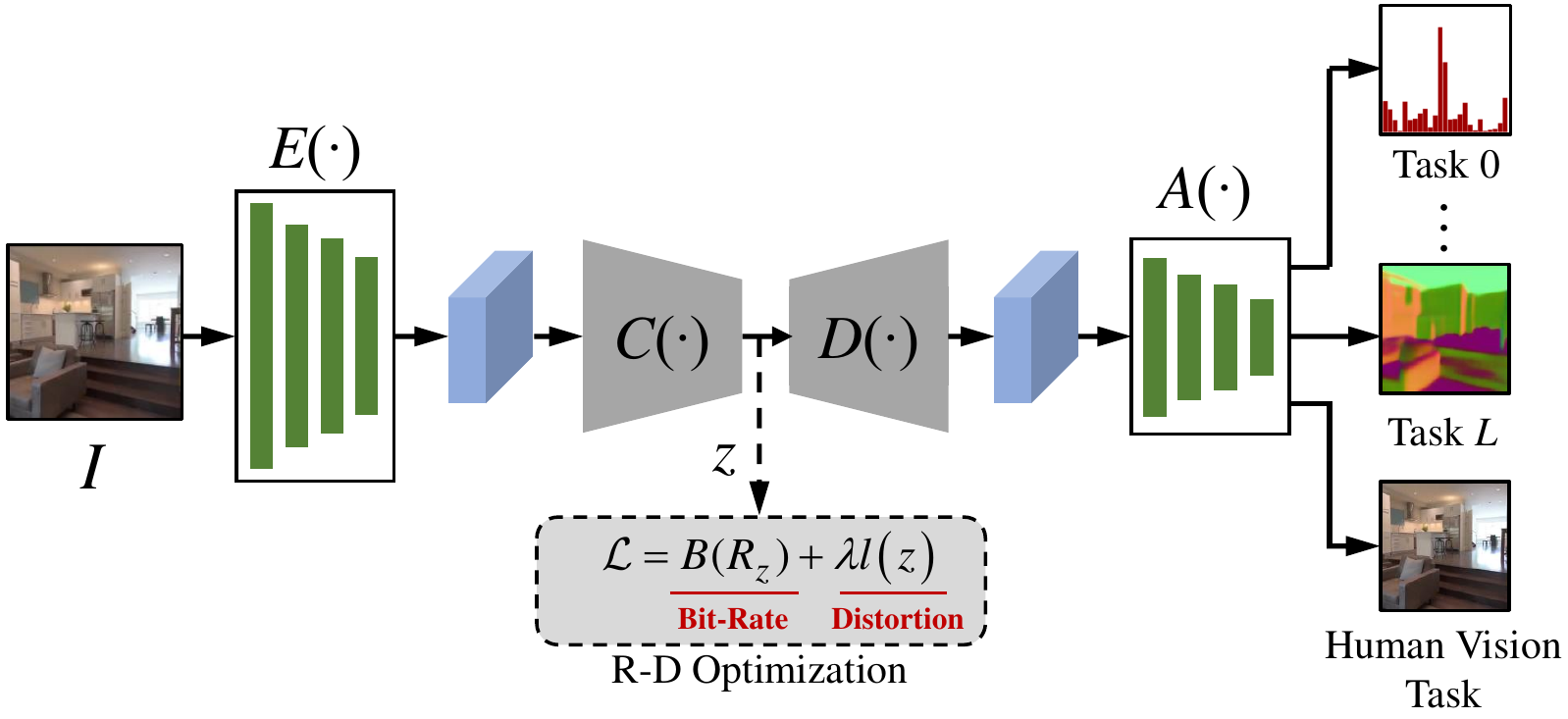}
		}
		\subfigure[Intermediate Feature Optimization]{
			\includegraphics[width=1\linewidth]{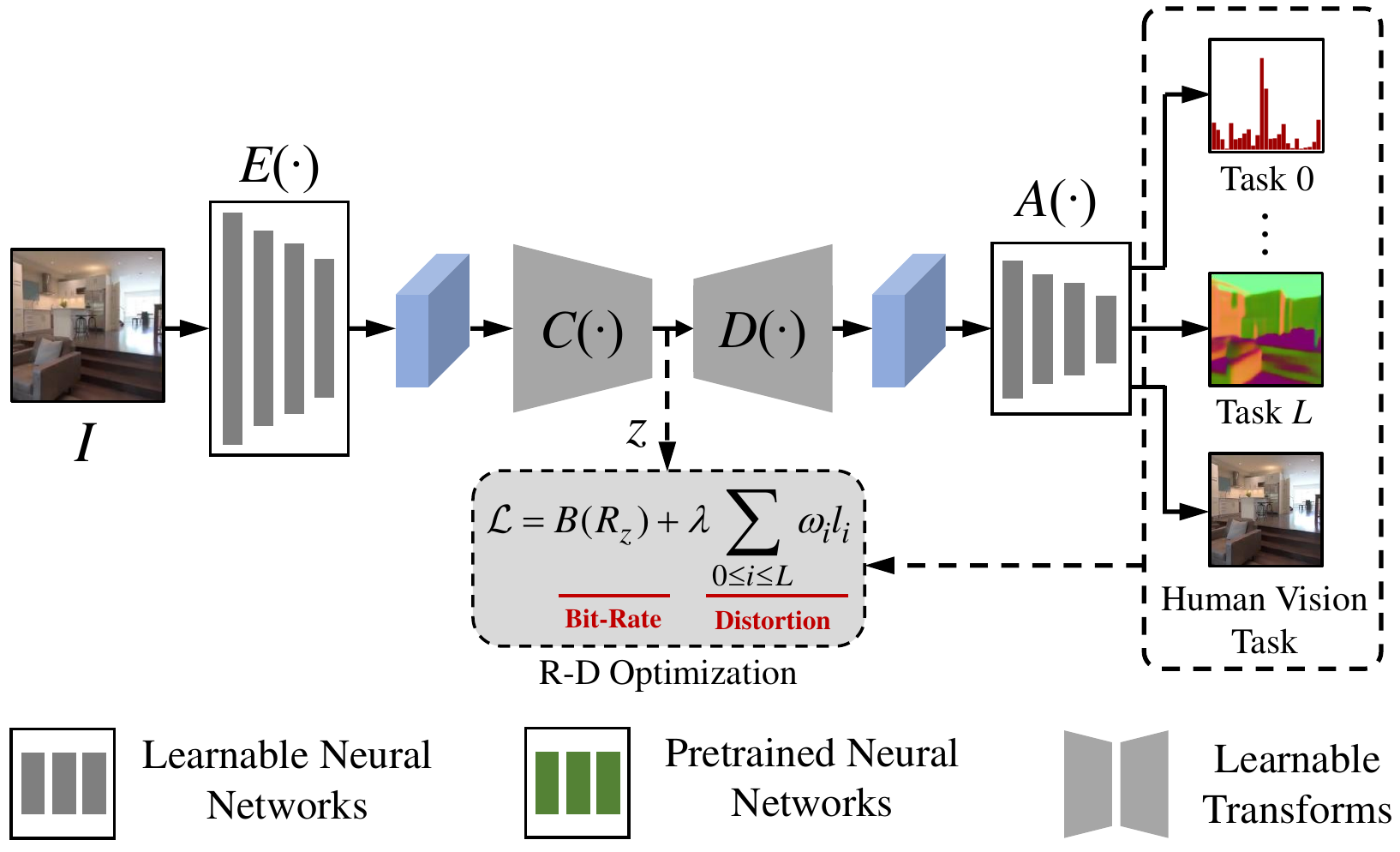}
		}
		\caption{The frameworks of intermediate feature compression and optimization.}
		\label{fig:feature_compression}
	\end{figure}

	\begin{figure*}[t]
	\centering
	\subfigure[Adaptive Mode]{
		\includegraphics[width=0.31\linewidth]{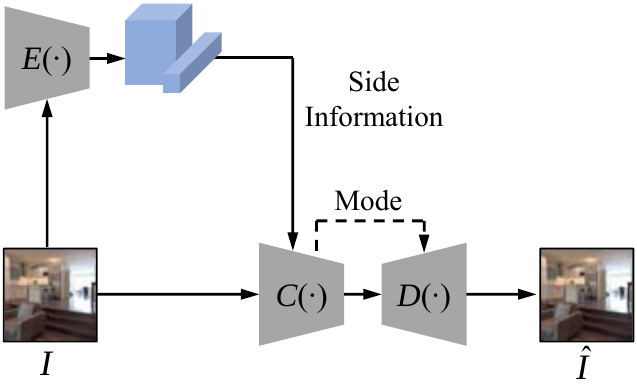}
	}
	\subfigure[Generative Coding]{
		\includegraphics[width=0.31\linewidth]{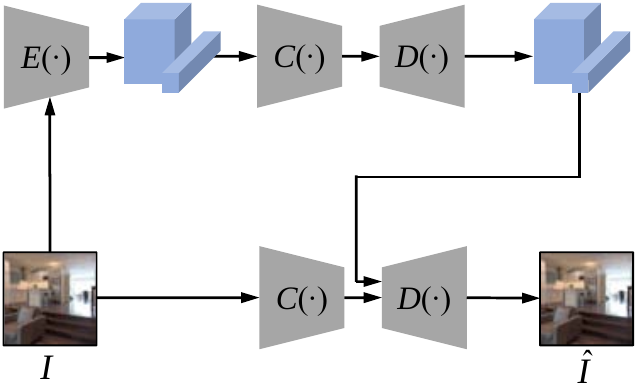}
	}
	\subfigure[Layered Coding]{
		\includegraphics[width=0.31\linewidth]{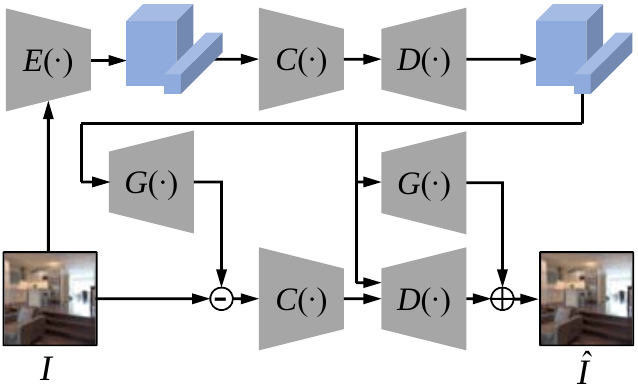}
	}
	\caption{
	Three frameworks of feature assisted coding.
	This category only considers the need for human vision.
	}
	\label{fig:feature_assisted_coding}
	\end{figure*}
		
	\input{summary}
	
	\subsection{Progress Survey}
	
	Based on the model's input, output, supporting tasks, and optimized terms, we categorize existing VCM methods into five classes: machine vision targeted codec, intermediate feature compression, intermediate feature optimization, features assisted coding, and scalable coding. 
	We will discuss the existing methods of the five classes in detail in the subsequent sections.
	A comprehensive summary of previous works is given in Table~\ref{tab:vcm}.
	
	\subsubsection{Machine Vision Targeted Codec}
	
	The first category of methods is a natural evolution of existing image/video codecs.
	As shown in Fig.~\ref{fig:machine_targeted_codec}, the final outputs of the models are still images/videos.
	Differently, the compressed images/videos do not serve humans but are fed into machines to support machine vision tasks.
	These methods do not get involved with the optimization of feature extractor $E(\cdot | \theta_E )$, analytics model $A(\cdot | \theta_A)$, and feature predictor $G(\cdot \theta_G)$.
	In Eq.~\eqref{eq:vcm1}, only one bit-stream is extracted from the image, and task-driven constraints are enforced on the reconstructed images or directly on the bit-stream.
	
	There are three directions in this category based on different ways to make the reconstructed images/videos include the information that benefits machine vision tasks, including \textit{side information guidance}, \textit{machine vision constraint}, and \textit{semantic information preservation}.
	The first direction -- \textit{side information guidance} -- is to detect the side information related to the analytics performance at first and then utilize the side information to adjust the coding configurations at the encoder and decoder sides.
	For example, in~\cite{Huang2021icme}, the region of interest for machine is detected based on the degree of importance for each coding tree unit in visual analysis, which is injected into a novel CTU-level bit allocation model.
	In~\cite{Choi2020eccv}, the task-specific quantization tables are learned via learning a differentiable loss function to approximates bit-rates.
	
	The second direction, \textit{i.e.} \textit{machine vision constraint}, adopts the loss functions that target machine vision optimization to train end-to-end learned codecs.
	In~\cite{Suzuki2019icip,Le2021icassp,Le2021icme,chamain2021end,Hou2020cvpr}, the task-driven losses are adopted.
	Namely, in these works, the downstream deep networks for machine vision are connected to the outputs of the codecs, 
	and these two parts are trained jointly.
	Besides, in~\cite{Le2021icme}, the perceptual loss is additionally utilized in the R-D optimization function of an inference-time content-adaptive finetuning scheme leading to higher compression efficiency for machine consumption.
	In~\cite{Yang2020acmmm}, the maximum mean discrepancy is adopted to align feature distributions, which results in preserving more consistent perception in the feature domain and better recognition of pre-trained machine vision models.
	
	The third direction~\cite{orzan2007structure}, \textit{i.e.} \textit{semantic information preservation}, constrains the encoded/decoded bit-streams to have the capacity of semantic preservation via connecting a classifier to the decoded bit-stream that predicts the semantic labels for machine vision tasks.
	
	To summarize, machine vision targeted codec still outputs images/videos naturally perceived by humans, which are more friendly to the successive machine vision tasks.
	However, they still need a whole analytics model at the cloud to perform the machine vision tasks, which in fact increases the whole burden at the front-end and cloud side.

	\subsubsection{Intermediate Feature Compression}
	The second category targets compressing features.
	Instead of compressing dedicated features for given tasks, as shown in Fig.~\ref{fig:feature_compression}~(a), the new methods aim to compress the deep intermediate features, which are expected to be more expressive and compact for the successive machine vision applications.
	This feature compression paradigm also allows the ``front-cloud'' collaborative processing.
	Namely, a part of inference calculation can be placed at the front end.
	From the perspective of Eq.~\eqref{eq:vcm1}, one layer of intermediate feature is compressed and the task performance is defined as the fidelity between the decoded feature and original one, \textit{i.e.} without a direct connection to the end-task performance.

	Chen \textit{et al.}~\cite{chen2019toward,chen2019acmmm} made the first attempt and presented lossy/lossless compression frameworks based on High Efficiency Video Coding (HEVC) as well as evaluation metrics for intermediate deep feature compression.
	The follow-up works further improve the coding efficiency via optimizing sub-modules in the framework, most of which focus on removing inter-channel redundancy.
	In~\cite{Suzuki2020icip}, Suzuki \textit{et al.} proposed a new feature arrangement method that regards the deep intermediate feature as videos and compressed the videos to make full use of spatio-temporal correlation.
	In~\cite{Ikusan2021icme}, Ikusan \textit{et al.} also compressed deep features from the perspective of video compression, especially distinguishing and making use of key frames.
	A selection strategy is developed to reduce the feature redundancy and the selected features are then compressed via video encoder.
	Then, an R-D optimization targeted for computer vision tasks is integrated into the codecs.
	In~\cite{choi2020icassp}, Choi \textit{et al.} also selected a subset of channels of the feature tensor to be compressed and introduced a novel back-and-forth prediction method to infer the original features of shallow layers based on compressed deep layers.

	In~\cite{chen2020icip}, Chen~\textit{et al.} proposed three new modes to repack features and explored two modes in Pre-Quantization modules to further improve the fidelity maintenance capacity.
	In~\cite{xing2020dcc}, Xing \textit{et al.} adopted logarithmic quantization and HEVC inter encoding to compress 3D CNN features for action recognition.
	In~\cite{hu2020vcip}, the different channels' contributions to the inference result are studied and a  channel-wise bit allocation is developed. The model takes a two-pass step.
	In the first step, the channel sensitivity is estimated while in the second step, bits are allocated based on the estimated sensitivity.
	Ulhaq and Baji\'c~\cite{Ulhaq2021icassp} explored the motion relationship between the corresponding feature tensors and concluded that the feature's motion is approximately equivalent to the scaled version of the input motion.
	
	To summarize, this category adopts existing codecs to compress deep intermediate features.
	However, these codecs are originally designed to deal with the image/video signals, which might be non-optimal to estimate the bit-rate of features and turn the features into compact representations.
	Furthermore, as the feature fidelity is very hard to accurately define and the model at the decoder side is not jointly optimized, the compression efficiency is not fully optimized.

	\subsubsection{Intermediate Feature Optimization}
	Beyond intermediate feature compression, this category also aims to compress deep intermediate features,
	but chooses to optimize the whole compression framework jointly with the successive machine vision tasks instead of feature fidelity, as shown in Fig.~\ref{fig:feature_compression}~(b).
	Alvar and Baji\'c~\cite{Alvar2020icassp} made the first effort to develop a loss function to measure a feature's compressibility that constrains the training process of multi-task models.
	Similarly, Singh \textit{et al.}~\cite{singh2020end} developed a penalty to train the network in an end-to-end manner for balancing the expressiveness and compressibility of deep features.
	Shah and Raj~\cite{Shah2020icassp} proposed an Annealed Representation Contraction method that obtains small-scale networks~(features) via iteratively tuning the shrunk models with network layer contraction and annealed labels.
	Zhang \textit{et al.}~\cite{zhang2021icme} developed a multi-scale feature compression method that is jointly optimized in a Mask RCNN trained with mask, box, and class-related losses.
	In~\cite{Alvar2020icassp,Alvar2021tip}, the task distortion is approximated as convex surfaces, which helps derive a closed-form bit allocation solution for both single-task and multi-task systems, and analytical characterization of the full Pareto set.
	
	In summary, it is beneficial to collaborative intelligence with the help of both intermediate feature compression/optimization.
	However, the methods in this category cannot provide full pixel reconstruction, \textit{i.e.} reconstructed images/videos.
	Besides, the compressed and transmitted features still rely on joint optimization with the successive machine vision tasks, which sets barriers for the generalization capacity of the features.

	\subsubsection{Feature Assisted Coding}
	The fourth category explores collaborative operations between video and feature streams to optimize the coding efficiency from the perspective of human vision, called feature assisted coding.
	As shown in Fig.~\ref{fig:feature_assisted_coding}, the side information or semantic features are first extracted, and these features are used to facilitate the full-pixel image/video reconstruction. In formulation, we have
	\begin{align}
		& \widetilde{B} \left( \left\{ R_{F_{i}} \right\}_{0 \le i \le L} \right)  = B \left( {R_{F_{0}}} \right) + \sum_{1 \le i \le L} {\mathop {\min }\limits_{0 \le j < i} } \left\{ { B \left( {\tiny }{{{{R_{F_{i \to j}}}}}} \right)} \right\}, \nonumber \\
		& {R_{{F_0}}}  = C\left( {{F_0}|{\theta _C}} \right), \label{eq:f_0}  \\
		& {R_{{F_{i \to j}}}} = C\left( {F_i} - {\rm{{G}}}\left( {{\widehat{F}_j},i|{\theta _G}} \right)\right | {\theta _C}), \text{ }{\rm{ for }} \text{ } i \ne 0, \nonumber 
	\end{align}
	where $B(\cdot)$ measures the bit-rate.
	Namely that, the most abstract feature ($F_0$) is first extracted while the task dependency is fully squeezed out via feature prediction $G(\cdot | \theta )$.
	
	There are three directions in this category: \textit{adaptive mode}, \textit{generative coding}, and \textit{layered coding}.
	The first direction -- \textit{adaptive mode} -- is shown in Fig.~\ref{fig:feature_assisted_coding}~(a).
	The side information is adopted to control the operations of the encoder and decoder.
	In~\cite{xia2020icme}, an object segmentation network is utilized to separate the (non-)object masks,
	and two compression networks are used to compress object regions and background ones, respectively.
	In~\cite{Chen2019icassp}, the pixel-level texture regions are inferred by semantic segmentation with the help of motion cues injected into codecs.
	In~\cite{li2019adacompress}, the coding process-related configuration and parameters, \textit{i.e.} the JPEG configuration, are generated with an agent to infer the compression level adaptively based on extracted features and deep network backbones.
	
	The second direction \textit{generative coding}, as shown in Fig.~\ref{fig:feature_assisted_coding}~(b), compresses and transmits the extracted features to form a single feature stream. 
	On the decoder side, the transmitted feature will help the decoding of the image/video stream in a generative way.
	In~\cite{Chun_2019_CVPR_Workshops}, gray-scale and color hints are extracted and compressed via BPG with an adaptive quantization parameter (QP). 
	At the decoder side, these two parts are processed by the artifact reduction network and colorization network.
	In~\cite{Prabhakar2021dcc}, the body pose and face mesh are detected at the encoder side and reconstructed into animated puppets at the decoder side to support video reconstruction.
	In~\cite{Kim2020icassp}, a generative decoder is adopted to map key frames as well as soft edges of non-key frames into the whole reconstructed frames.
	In~\cite{chang2019layered}, the edges are extracted to form the feature stream, which facilitates the image reconstruction at the decoder side.
	
	The third direction \textit{layered coding}~\cite{Akbari2019icassp}, as shown in Fig.~\ref{fig:feature_assisted_coding}~(c), further introduces the prediction mechanism to remove the redundancy between feature and image streams.
	The segmentation map plays the role of the base layer.
	A compact image then acts as the first enhancement layer.
	These two layers are used to form a coarse reconstruction of the image.
	The residue layer, namely, the difference between the input and the coarse reconstruction, acts as another enhancement layer.
	
	In summary, feature assisted coding introduces a scalable mechanism to improve coding efficiency. However, these methods do not consider supporting machine vision tasks, which will be investigated in the next sub-section.
	
	\begin{figure}[t]
		\centering
		\subfigure[Two-Stream Scalable Coding]{
			\includegraphics[width=1\linewidth]{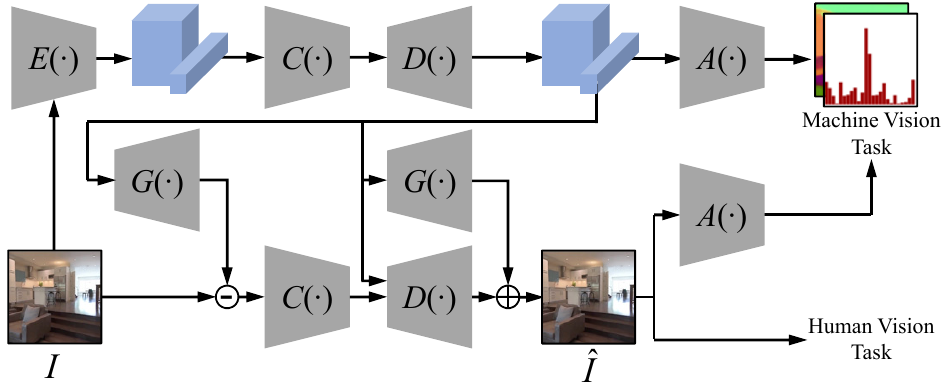}
		}
		\subfigure[Multi-Stream Scalable Coding]{
			\includegraphics[width=1\linewidth]{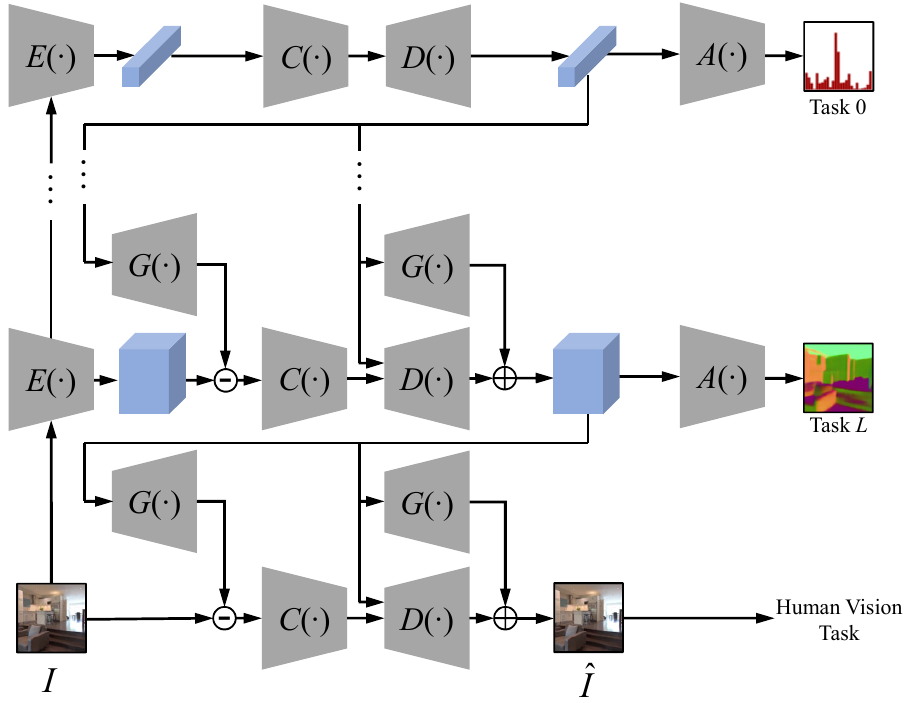}
		}
		\caption{
		The frameworks of scalable coding.
		Beyond feature assisted coding, this category tries to fulfill the need of both human and machine vision.
		}
		\label{fig:scalable_coding}
	\end{figure}
	
	\subsubsection{Scalable Coding}
	
	This category also compresses and transmits both feature and image/video streams to serve both human and machine visions via the same route of Eq.~\eqref{eq:f_0}.
	There are two directions in this category.

	The first direction \textit{two-stream scalable coding}, as shown in Fig.~\ref{fig:scalable_coding}~(a), takes the same architecture as \textit{layered coding} in Fig.~\ref{fig:feature_assisted_coding}~(c).
	However, differently, besides supporting the image/video reconstruction, 
	the feature stream transmitted by the methods here needs to support the machine vision tasks.
	The works in ~\cite{hu2020icme,yang2021tmm,wang2021icassp,wang2021tmm} focus on the analysis and reconstruction of face images via the two-stream structures.
	In~\cite{hu2020icme,yang2021tmm}, quantized edges and color clues are used to generate reconstruction images via the generative model at different bit-rates that support facial landmark detection and image reconstruction.
	In~\cite{wang2019scalableface,wang2021tmm,wang2021icassp}, in a scalable coding framework, the deep learning feature acts as the base layer to support face recognition, and the enhancement layer learns to reconstruct the full-pixel images.
	In~\cite{change2021icme}, the semantic prior is modeled and transmitted to facilitate the conceptual coding scheme towards extremely low bit-rate image compression, where the reconstructed images serve facial landmark detection.
	In~\cite{Hoang2020cvprw}, the semantic segmentation map and reconstructed image compensate for each other to improve the semantic and visual quality.
	In~\cite{xia2020emerging}, the learned motion pattern, \textit{i.e.} key points as well as the related trajectories, is transmitted for action recognition and video frame reconstruction.
	
	The second direction \textit{multi-stream scalable coding,} as shown in Fig.~\ref{fig:scalable_coding}~(b), adopts more than one stream to transmit features with a feature salable prediction mechanism.
	In~\cite{choi2021arxiv}, side, base, and enhancement streams are transmitted together, where the side stream is used for entropy estimation and control, the base stream serves machine vision tasks, while the enhancement stream helps compensate for full-pixel image reconstruction.
	In~\cite{yan2020icip}, a feature compression approach is conducted on the intermediate
	representations where different layers serve the tasks requiring different grained semantic information.
	The work in~\cite{liu2021ijcv} develops a lifting structure based on a trainable and revertible transform that decomposes the image into different bands to support different machine vision tasks.
	
	In summary, with the scalable mechanism and feature stream, these methods can support both human and machine vision in an efficient and flexible way.
	However, some critical issues in intelligent collaborative analytics of massive data and diverse tasks/applications, \textit{e.g.} the transferability among different tasks, the generality of the extracted features, \textit{etc.}, are still not explored.
	
	\begin{figure*}[t]
		\centering
		\includegraphics[width=1\linewidth]{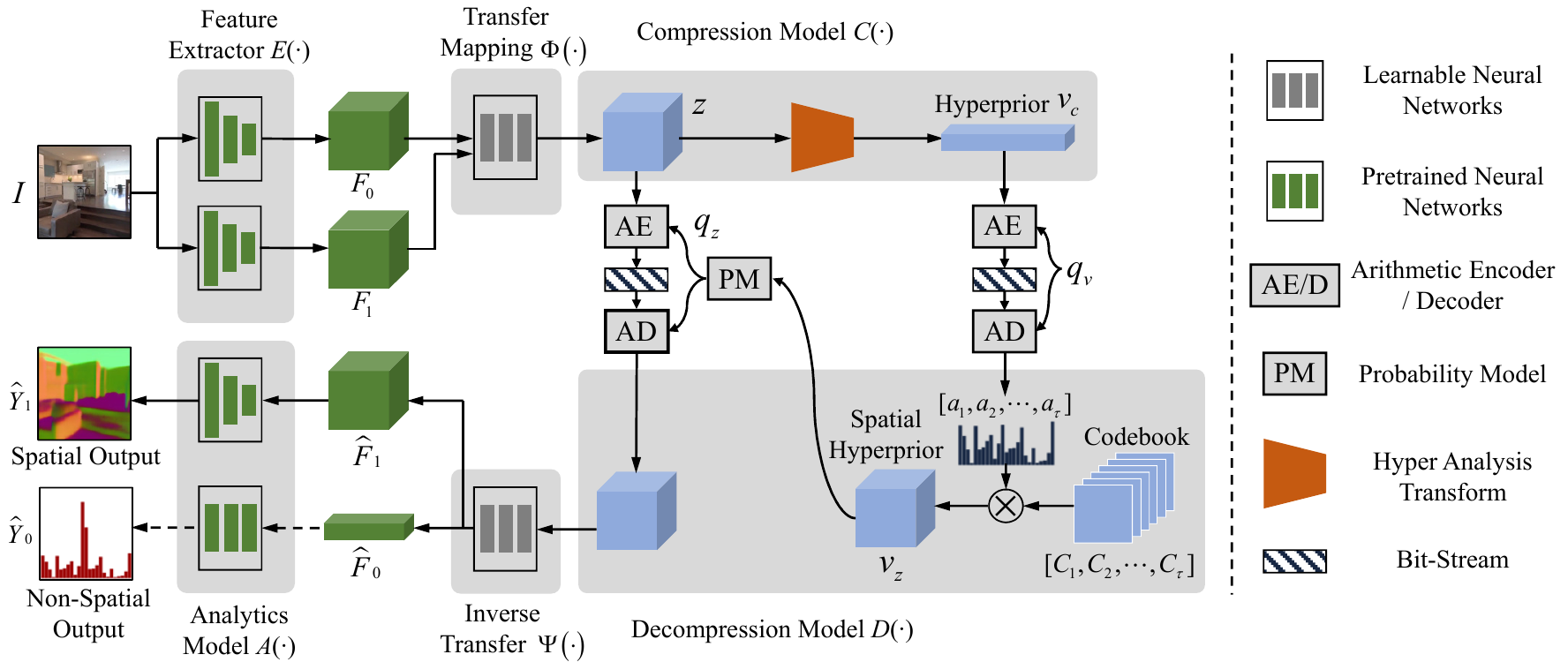}
		\caption{
			The visual representation compression in analytics taxonomy for deep features of multiple tasks.
			With a pre-trained neural network (in {\color{green}green}), a feature tensor is extracted and compressed by the proposed model (in {\color{blue}blue}).
			The reconstructed feature tensor is processed by the rest layers of the pre-trained network to produce analytics results. Dashed lines illustrate the processing of the feature vectors without the spatial dimension.}
		\label{fig:codebook}
	\end{figure*}
	
	\section{Revisit Visual Representation Compression in Analytics Taxonomy}
	\label{sec:analytics}
	
	\subsection{Formulation of Compressive Analytics Taxonomy}
	The previous paradigms still face issues when dealing with massive data and diverse kinds of tasks, \textit{e.g.} Taskonomy~\cite{zamir2018taskonomy}.
	\begin{itemize}
		\item Transmitting the multi-task features one-by-one leads to additional redundancies as different tasks inevitably have semantic gaps.
		\item Each feature is deeply coupled with the corresponding task, therefore is hard to be generalized to handle unseen tasks.
		\item Compressing and transmitting multi-task features one-by-one in a scalable way results in greater latency and complexity.
	\end{itemize}
	Therefore, we propose to investigate the relationship among different feature representations from a compression perspective, and seek to construct a combined compact feature serving a bundle of tasks, which can also be generalized to deal with unseen tasks. More specifically, we have a reformulated VCM problem:
	\begin{align}
		\label{eq:vcm2}
		\footnotesize
		\mathop {\text{argmax}  }\limits_{ \widehat{\Theta} } & \sum\limits_{0 \le i \le L} {{\omega _i}{l_i}}, \text{ \textit{s.t.} } \sum\limits_{0 \le i \le L} {{\omega _i}}  = 1, \\
		& B\left( R_{{z}} \right)
		\le {S_T}, \nonumber
	\end{align}
	where a transfer function $\Phi \left( \cdot | \theta_\Phi \right)$ and an inverse-transfer function $ \Psi \left( \cdot | \theta_{\Psi} \right)$ with  $\theta_\Phi$  and $ \theta_{\Psi} $ as their parameters, are adopted to map multi-task features into and from a combined feature $z$, respectively.
	The related important elements are defined as follows,
	\begin{align}
		{R_{{z}}} & = C\left( { {\Phi\left( \left\{ F_{i} \right\}_{0 \le i \le L} | \theta_\Phi \right)  } |{\theta _C}} \right), \label{eq:z} \\
		{\widehat F_i} & = \left[\Psi \left( D\left( {R_{z}|{\theta _D}} \right) | \theta_{\Psi} \right)\right]_i, \label{eq:back} \\
		\widehat{\Theta} & = \left\{ {{\theta _E}, {\theta _C},{\theta _D},{\theta_G},{\theta_A},  	{\theta_\Phi}, {\theta_\Psi} } \right\}, \label{eq:para}
	\end{align}
	where $\left[ \cdot \right]_i$ denote to select the $i$-th element from the set.
	Eq.~\eqref{eq:z}-\eqref{eq:para} show the R-D optimization route for multiple tasks where a combined compact feature $z$ is first derived via $\Phi \left( \cdot | \theta_\Phi \right)$ and different tasks are jointly supported by this unified representation via $ \Psi \left( \cdot | \theta_{\Psi} \right)$.
	
	In this paradigm, when handling multiple tasks, the scalable feature prediction is merged into decoding and transfer mapping, which further improves coding efficiency and reduces the complexity as well as system delay. Furthermore, as the new feature $z$ contains information of multiple tasks rather than bundled with the given task, the framework has the potential to be well generalized to handle unseen tasks.
	
	\subsection{Codebook-Hyperprior Model for Deep Feature Compression}
	
	
	In order to solve the optimization function in Eq.~\eqref{eq:vcm2},
	we first design a trainable compression framework targeting to compress deep features, illustrated in Fig.~\ref{fig:codebook}, to estimate and reduce the information entropy of each deep feature representation $F_i$.
	Then, in the next subsection, we consider handling deep features extracted from multiple tasks.
	
	\subsubsection{Entropy Modeling}
	We transform $F_i$ into a compact $z$, whose probability distribution is tractable and can be compactly compressed into bit-stream.
	As the mainstream neural networks do not apply any constraint on its generated $F_i$, the probability distribution of $F_i$ is usually unknown and it is intractable to estimate the entropy of $F_i$.
	Therefore, we apply a transform to $F_i$ and obtain a functionally equivalent representation $z$. 
	The transform makes $z$ have the desired signal structure. Thus, its probability distribution is tractable.
	Hence, we can estimate the entropy of $F_i$ via calculating the entropy of the structured representation $z$.
	
	Specifically, $z$ fits well with a successive entropy estimation process: elements of $z$ have been quantized into integers. 
	The value of each element $z_k$ belongs to a finite set $\mathbb{S}=\{t_{min}, \cdots, -1, 0, 1, \cdots, t_{max}\}$, 
	and thus $z$ is sampled from the finite space $\mathbb{S}^K$, where $K$ denotes the dimension of $z$.
	Given a probability distribution in the finite space, the information entropy can be calculated as,
	\begin{equation}
		H(z) = \sum_{z \in \mathbb{S}^K} -p(z) \log p(z),
	\end{equation}
	which is the lower bound of the average bit-rate to encode $z$. 
	
	\subsubsection{Hyperprior Model}
	As it is usually intractable to estimate $p_z$, we adopt a parametric probability model $q_z$ to estimate the probability distribution of $z$ during the encoding.
	The actual bit-rate to encode $z$ with the probability $p_z$ under an estimated entropy model $q_z$ equals to the cross-entropy~\cite{cover1999elements} of $p$ and $q$, as,
	\begin{equation}
		\begin{split}
			\label{eq:hpq}
			H(p, q) = \mathbb{E}_{p} [-\log q] = H(p) + D_{KL}(p||q).
		\end{split}
	\end{equation}
	It has been shown in Eq.~(\ref{eq:hpq}) that $H(p) \leq H(p,q)$, where the equality is achieved when $D_{KL}(p||q) = 0$, \textit{i.e.} when the probability model $q$ estimates $p$ perfectly.

	Ball{\'e} \textit{et al.}~\cite{balle2018variational} proposed to extract and encoded a hyperprior from an image representation for more accurate entropy estimation and constraint.
	The hyperprior is a part of the signal/feature that is first transmitted from the encoder side to the decoder side for estimating the entropy of other signal/feature parts.
	More specifically, the hyperprior is often of a lower resolution and used to estimate the probability distribution of the corresponding image representation.
	A hierarchical structure of a hyperprior can further improve the accuracy of the probability estimation for image representation~\cite{hu2020coarse}.
	
	However, feature representations $F_i$ and $z$ are not image-level dense pixel signals.
	They serve machine vision tasks and do not include the information of image appearances.
	Although their extracted features might take the form of tensor (not vector) and have spatial dimensions, these features in fact can be embedded into very low-dimensional space, which does not have the spatial signal structure.
	As the image compression oriented hyperprior model relies on the hierarchical spatial structure of images, it becomes less effective to model the signal structure of $F_i$ and $z$, resulting in a gap between $p$ and $q$ and making the entropy estimation less effective.

	\subsubsection{Codebook-Hyperprior Model}
	To reduce the gap, we assume that the extracted feature representations from the neural network can be embedded into a very low-dimensional manifold.
	Each observed instance can be regarded as a point sampled from that the low-dimensional subspace, and the perturbation is independently distributed conditioned on the coordinates that expand the space.
	This assumption naturally leads to the proposed low-dimensional hyperprior model.
	
	The main idea is that, we adopt the hyperprior vector without the spatial dimensions in the encoding process to capture the intrinsic signal structure of $F_i$ and $z$, but transform the hyperprior vector into the hyperprior tensor with the spatial dimensions in the decoding process to augment the hyperprior's modeling capacity.
	To estimate the probability distribution of $z$, a hyperprior $v_c$ is extracted from $z$ via a hyper analysis transform $f_{Ha}(\cdot)$ as, namely, $v_c = f_{Ha}(z)$.
	The estimation of probability $p(z)$ can be divided into $p(z) = p(z,v_c) = p(v_c)p(z|v_c)$.
	Then, we apply a global pooling operation to reduce the spatial dimensions of $z$, producing $v_c$ in the vector form. Note that $v_c$ is also quantized to integers.
	We further assume that each element in $v_c$ follows a zero-mean Gaussian distribution $\mathcal{N}(0,\sigma_j)$. 
	Conditioned on $v_c$, each element $z_k$ in $z$ is conditionally independently distributed.

	The entropy of $v_c$ is estimated by tuning the parameter $\sigma_j$.
	We model $q_{z_k|v_c}$ with a Gaussian distribution $q_{z_k|v_c} \sim \mathcal{N}(\mu_k=f(v_c;\theta_{f}), \sigma_k=g(v_c;\theta_{g}))$, where the mean and scale are generated via a function taking $v_c$ as the input.
	To achieve this, we decode $n$ sequences of coefficients from $v_c$.
	Each sequence $\mathbf{A}^s = \left(a_1^s, a_2^s, \cdots, a_{\tau}^s\right), s \in [1,n]$ indicates a linear combination of the spatial bases, defined by a codebook, in the form of $\{C_1, C_2, \cdots, C_{\tau}\}$.
	With the codebook and the sequences of coefficients $\{\mathbf{A}\}_n$, we generate the spatial hyperprior $\hat{\mathbf{Z}}$ as,
	\begin{equation}
		\begin{split}
			\hat{Z}_l &= a_1^s C_1 + a_2^s C_2 + \cdots + a_{\tau}^s C_{\tau} \text{ , for  } s=1,2, \cdots, n,\\
			v_z &= (\hat{Z}_1, \hat{Z}_2, \cdots \hat{Z}_n).
		\end{split}
	\end{equation}
	We employ a prediction sub-network to estimate $\mu_k=f(v_c;\theta_{f}), \sigma_k=g(v_c;\theta_{g})$ from $v_z$.
	By learning the parameters of the sub-network, $\theta_{f}$ and $\theta_{g}$ are estimated to provide an accurate estimation $q(z|v_c)$ for $p(z|v_c)$.
	%
	%
	The spatial dimensions of the codebook $\{C_1, C_2, \cdots, C_{\tau}\}$ are fixed, and therefore it requires a re-sampling to deal with the inputs of different resolutions.
	
	The proposed model is also general and flexible to support the deep features without spatial dimensions,
	\textit{i.e.} feature vectors. 
	This can be achieved by directly producing the vector-form probability parameters $\mu_k=f(v_c;\theta_{f}), \sigma_k=g(v_c;\theta_{g})$ with $v_c$, via multi-layer perceptions.
	
	\begin{table*}[htbp]
	\footnotesize
	\centering
	\vspace{-5mm}
	\caption{Experimental results for the semantic segmentation task with various compression schemes. 
		Method IDFC is evaluated with different QPs, marked as IDFC~(QP) in the table. $\uparrow$ means higher performance, better result, and $\downarrow$ vice versa.
		The best results are denoted in bold.
	}
	\begin{tabular}{cccccc}
		\hline
		Method & Bit-Rate (bpp)$\downarrow$ & Cross Entropy$\downarrow$ & Acc.$\uparrow$ & Non-BG Acc.$\uparrow$ & mIoU$\uparrow$ \\
		\hline
		Original & /   & 0.74  & 91.64\% & 86.28\% & 27.65\% \\
		Control Group & /   & 0.61  & 92.31\% & 82.67\% & 27.07\% \\
		\hline
		IDFC~(qp=51)~\cite{chen2019toward} & 0.020 & 6.22  & 92.74\% & 15.62\% & 11.03\% \\
		IDFC~(qp=43)~\cite{chen2019toward} & 0.026 & 1.38  & \textbf{93.94}\% & 72.24\% & 28.41\% \\
		Hyperprior~\cite{chamain2021end} & 0.025 & 0.80  & 91.95\% & 79.64\% & 25.42\% \\
		Ours  & \textbf{0.013} & \textbf{0.77}  & 93.58\% & \textbf{81.35}\% & \textbf{29.35}\% \\
		\hline
	\end{tabular}%
	\label{tab:seg}%
\end{table*}%

\begin{figure*}[t]
	\centering
	\subfigure[IDFC (qp=51) vs. Ours]{
		\includegraphics[width=0.31\linewidth]{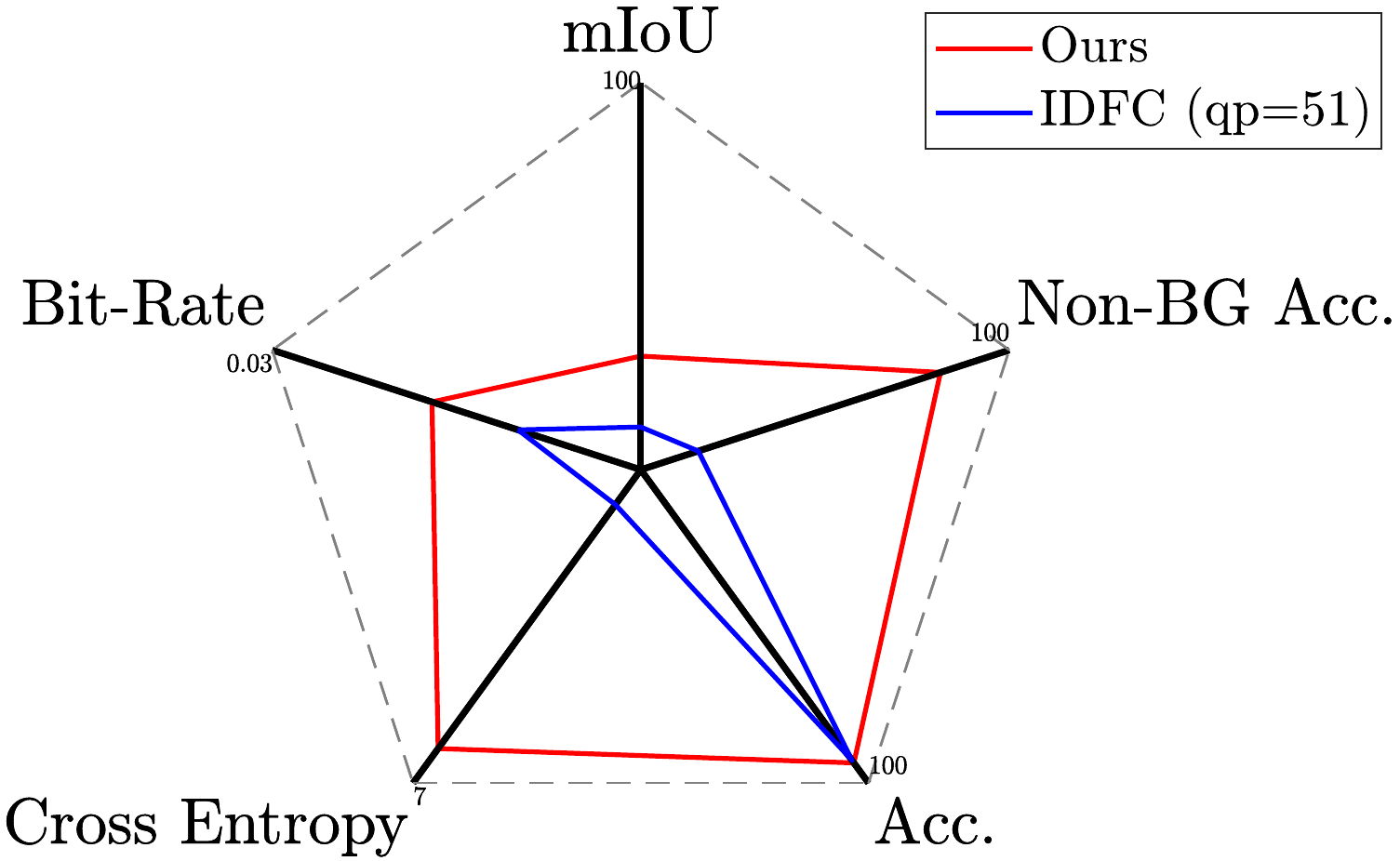}
	}
	\subfigure[IDFC (qp=43) vs. Ours]{
		\includegraphics[width=0.31\linewidth]{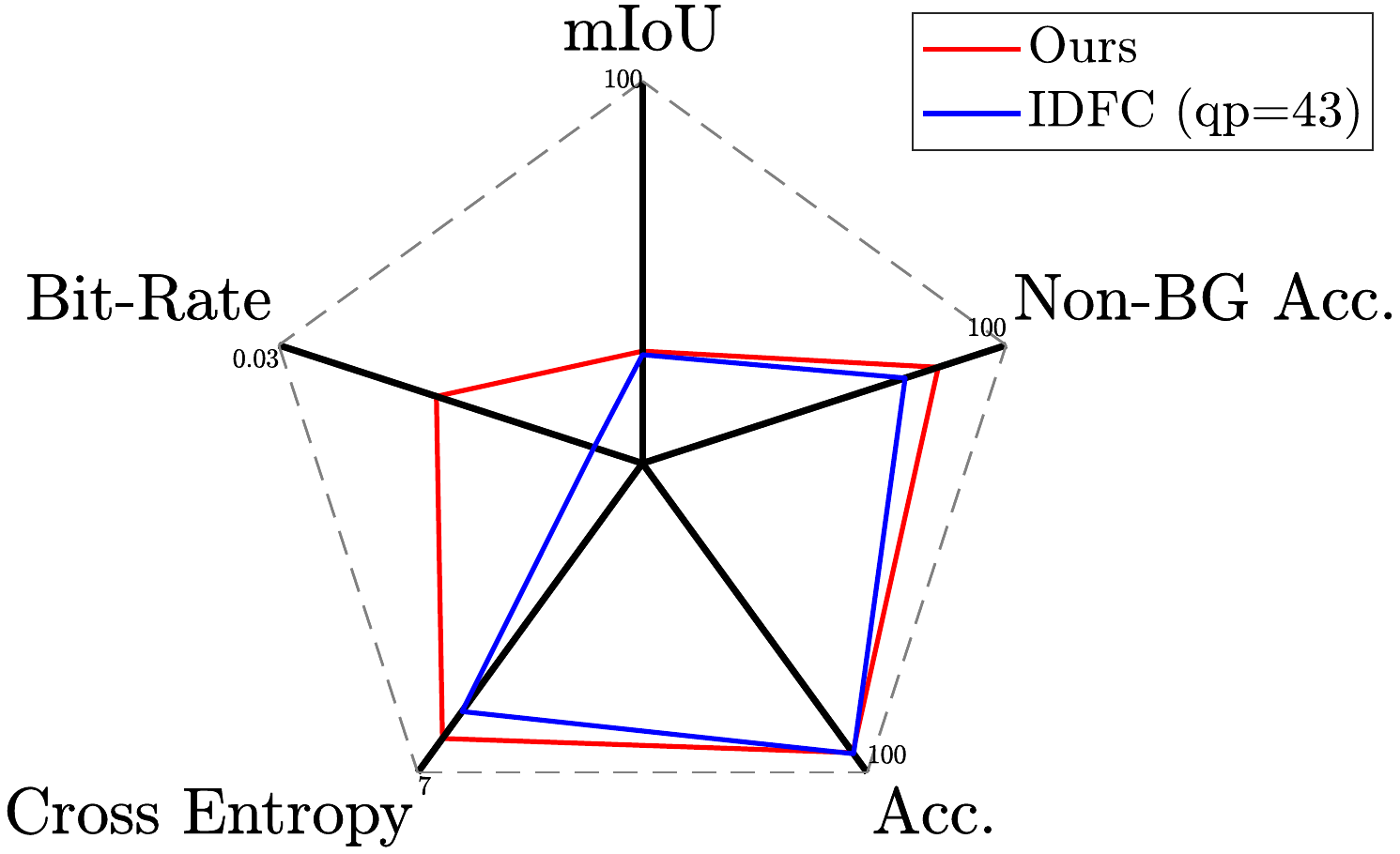}
	}
	\subfigure[Hyperprior vs. Ours]{
		\includegraphics[width=0.31\linewidth]{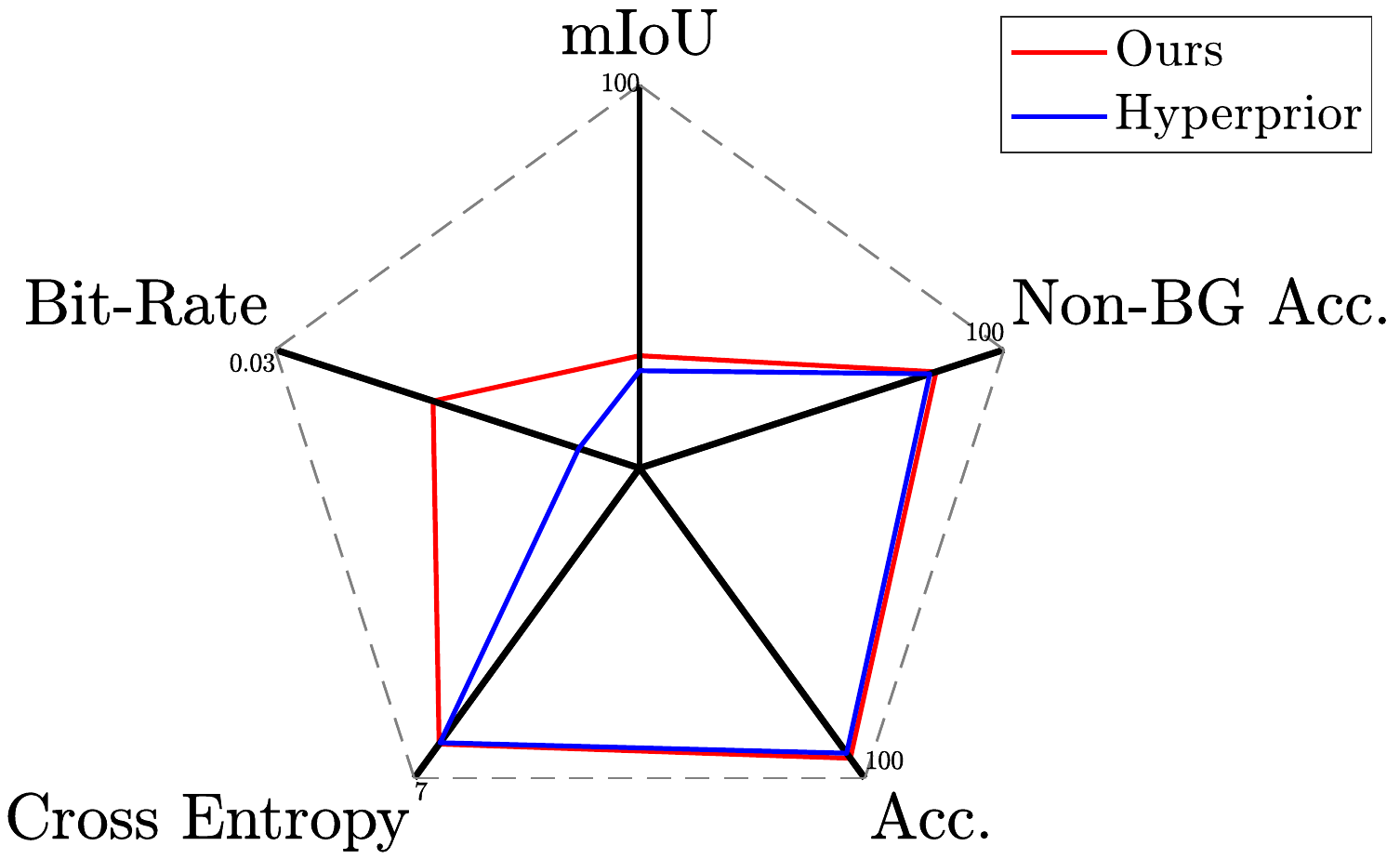}
	}
	\caption{The radar chart results on the semantic segmentation task with various compression schemes.
		The values of Bit-Rate ($x_1$) and Cross Entropy ($x_2$) are adjusted via $0.03-x_1$ and $7-x_2$ for better visibility, respectively.
	}
	\label{fig:radar_seg}
\end{figure*}

	\subsection{Transfer Mapping: Multi-Task Compression}
	
	\subsubsection{(Inverse-)Transfer Mapping}
	It has been shown in \cite{zamir2018taskonomy} that,
	there exist connections among feature representations of different tasks.
	Thus, if multiple tasks are supported as we mentioned in the problem formulation,
	the separate compression for each task may be less efficient due to the cross-task redundancy.
	Therefore, we propose the aggregation transformed compression scheme to generate the compressed representation for different tasks jointly.
	
	An example of the proposed aggregation transformed compression scheme is shown in Fig.~\ref{fig:codebook}.
	The illustrated structure compresses and aggregates the feature representations of two tasks into one bit-stream.
	Each representation is transformed with a sub-network. The transformed features are concatenated and compressed via a single compression model.
	The decompressed representation is then split via another set of convolutional layers, serving as the input of the rest of the pre-trained analytics network.

	\subsubsection{Rate-Distortion Loss}
	The aggregation transformed compression model is trained in two stages, 
	corresponding to the two application scenarios, including: 1) analytics oriented compression in a known set of tasks;
	and 2) out-of-set analytics, \textit{i.e.} handling unseen tasks.
	During the first training phase, the parameters of the compression model 
	and the multi-layer peripheral convolutions before the compression model for each task are tuned.
	Parameters of the pre-trained analytics models are fixed.
	The compression model learns to compress different forms of feature representations jointly. The parameters are trained with the joint R-D loss function as,
	\begin{equation}
		\begin{split}
			\mathcal{L} =  B\left( R_{{z}} \right) + \lambda \sum\limits_{0 \le i \le L} {{\omega _i}{l_i}},
		\end{split}
	\end{equation}
	where $\lambda$ is Lagrange multiplier to indicate the relative importance of bit-rate and distortion.
	%
	%
	%
	%

	\section{Experimental Results}
	\label{sec:exp}
	
	\subsection{Experimental Settings}
	\noindent \textbf{1) Dataset.} We conduct the experiments on the Taskonomy dataset~\cite{zamir2018taskonomy}, which contains approximately 4.5 million images, all labeled by 25 attributes, to support various machine vision tasks. 
	The following experiments are conducted on a subset.
	The subset is selected at random, while we control the numbers of images in the splits, \textit{i.e.} 51,316 images for training, 945 for validation, and 1,024 for testing. 
	Images in different splits of the data are captured in different buildings.
	Thus, the splits are diverse in content.
	We select a set of \textit{real-world} tasks for evaluation, \textit{i.e.} scene classification, object classification, semantic segmentation, surface normal estimation, reshading, and principle curvature estimation.
	The selected tasks include diverse categories, with which we evaluate both high-level semantics driven analytics and mid-level geometry related estimation.
	We follow the setting in~\cite{zamir2018taskonomy} to compare on $256\times256$ images, and calculate bits-per-pixel on that resolution.
	\vspace{1mm}
	
	\noindent \textbf{2) Pretrained Models.} The abundance of tasks link to one image provides the desired environment for our study. 
	We utilize the pre-trained models on the dataset, provided by the authors under the MIT License.
	All pre-trained models are hourglass encoder-decoder neural networks, as described in~\cite{zamir2018taskonomy}.
	\vspace{1mm}
	
	\noindent \textbf{3) Compared Methods}. We compare our method with the following baselines:
	\begin{itemize}
		\item \textit{Intermediate deep feature compression}~(\textit{IDFC})~\cite{chen2019toward};
		\item \textit{Hyperprior model}~\cite{balle2018variational} in the feature compression scheme~\cite{chamain2021end};
		\item A \textit{Control Group}. To avoid the potential bias due to the training procedure, we set up the \textit{Control Group} experiments, where a transform network with an identical structure to the compression model is trained, but no bit-rate constraint is applied.
		\item The \textit{Original} model. It refers to the results given by the originally provided hourglass-like networks in \cite{zamir2018taskonomy}.
	\end{itemize}
	Note that both the hyperprior model and the proposed scheme get involved in a training process.
	The models are initialized with the original weights released by~\cite{zamir2018taskonomy}. 
	%
	We select the model checkpoint with the lowest R-D cost and compare on the testing set.
	Both hyperprior and the proposed model are trained with $\mathcal{L}=R+\lambda \mathcal{L}_{CE}$.
	\vspace{1mm}
	
	\noindent \textbf{4) Evaluation Measures}. 
	We evaluate the performance of different methods for the tasks of semantic segmentation, scene classification, surface normal estimation, and reshading.
	For semantic segmentation, we adopt mean pixel-level accuracy~(Acc.), the accuracy of pixels in the non-background regions~(Non-BG Acc.), and mean IoU~(mIoU) by averaging the result among all 17 classes to provide a comprehensive performance evaluation.
	Besides, we measure the performance of the task scene classification in accuracy, surface normal estimation of indoor scenes in $L_1$ distance, and reshading of an indoor image in $L_1$ distance.
	Bit-per-pixel (bpp) is used to calculate the bit-rate usage for measuring the compactness in the compression.

	\begin{table}[t]
		\scriptsize
		\centering
		\caption{Evaluation of the plateau bit-rate for different tasks with the proposed method, IDFC and Hyperprior. We present the validation set performance (Val. Perf.) and the test set performance (Test Perf.) along with the related bit-rate. Performances of different tasks are evaluated in different metrics. $\uparrow$ means higher performance metric, better result, and $\downarrow$ vice versa.
		The best results are denoted in bold.}
		\begin{tabular}{cccccc}
			\hline
			Task  & Method & Val. Perf. & Val. bpp & Test Perf. & Test bpp \\
			\hline
			\multirow{5}[0]{*}{\tabincell{c}{Scene \\ Class$\uparrow$}} & Original & 70.02\% & /     & 67.48\% & / \\
			& Control Group & 75.66\% & /     & 62.70\% & / \\
			& IDFC  & 61.16\% & 0.0403  & 65.43\% & 0.0408  \\
			& {Hyperprior} & {67.48\%} & {0.0088} & {59.67\%} & {0.0102}\\
			& Ours  & 71.11\% & \textbf{0.0068}  & 59.47\% & \textbf{0.0069}  \\
			\hline
			\multicolumn{1}{c}{\multirow{5}[0]{*}{\tabincell{c}{Semantic \\ Seg.$\uparrow$} }} & Original & 18.37\% & /     & 27.65\% & / \\
			& Control Group & 18.85\% & /     & 27.07\% & / \\
			& IDFC  & 17.20\% & 0.0210 & 28.41\% & 0.0261 \\
			& {Hyperprior} & {19.02\%} & {0.0104} & {25.42\%} & {0.0250}\\
			& Ours  & 18.19\% & \textbf{0.0072} & 29.35\% & \textbf{0.0131} \\
			\hline
			\multirow{5}[0]{*}{\tabincell{c}{Surface \\ Normal$\downarrow$}} & Original & 0.0741 & /     & 0.1211 & / \\
			& Control Group & 0.0700 & /     & 0.1252 & / \\
			& IDFC  & 0.0753 & 0.0520 & 0.1281 & 0.0588 \\
			& {Hyperprior} & {0.0718} & {0.0402} & {0.1288} & {0.0454}\\
			& Ours  & 0.0721 & \textbf{0.0187} & 0.1299 & \textbf{0.0197} \\
			\hline
			\multirow{5}[0]{*}{Reshading$\downarrow$} & Original & 0.2209 & /     & 0.2836 & / \\
			& Control Group & 0.1687 & /     & 0.2343 & / \\
			& IDFC  & 0.2217 & 0.0830 & 0.2844 & 0.0959 \\
			& {Hyperprior} & {0.1844} & {0.0134} & {0.2382} & {0.0138}\\
			& Ours  & 0.1713 & \textbf{0.0130} & 0.2411 & \textbf{0.0134} \\
			\hline
		\end{tabular}%
		\label{tab:tasks}%
		\vspace{-2mm}
	\end{table}%
	
	\subsection{Efficacy of the Proposed Compression Scheme}
	
	We first evaluate the efficacy of the proposed codebook-hyperprior driven compression model for deep feature representations.
	The range of bit-rates we show in Table~\ref{tab:seg} is regarded as the \textit{plateau} bit-rate, where the compressed feature representation provides enough information to make the prediction accuracy comparable to the models without the bit-rate control.
	We also show the radar charts of our method compared to \textit{IDFC} and the primitive \textit{Hyperprior} model in Fig.~\ref{fig:radar_seg}, 
	which can better reflect the performance comparisons in all metrics.
	It is observed that the proposed method owns a larger area than \textit{IDFC} and \textit{Hyperprior}.
	The results in Fig.~\ref{fig:radar_seg} and Table~\ref{tab:seg} show that our model can better compress the deep features than existing methods~\cite{chen2019toward,chamain2021end}, as it consumes fewer bit-rates to reach a higher analytics performance in multiple metrics.

	\begin{figure}[t]
	\centering
	\subfigure[Scene Class]{
		\includegraphics[width=0.47\linewidth]{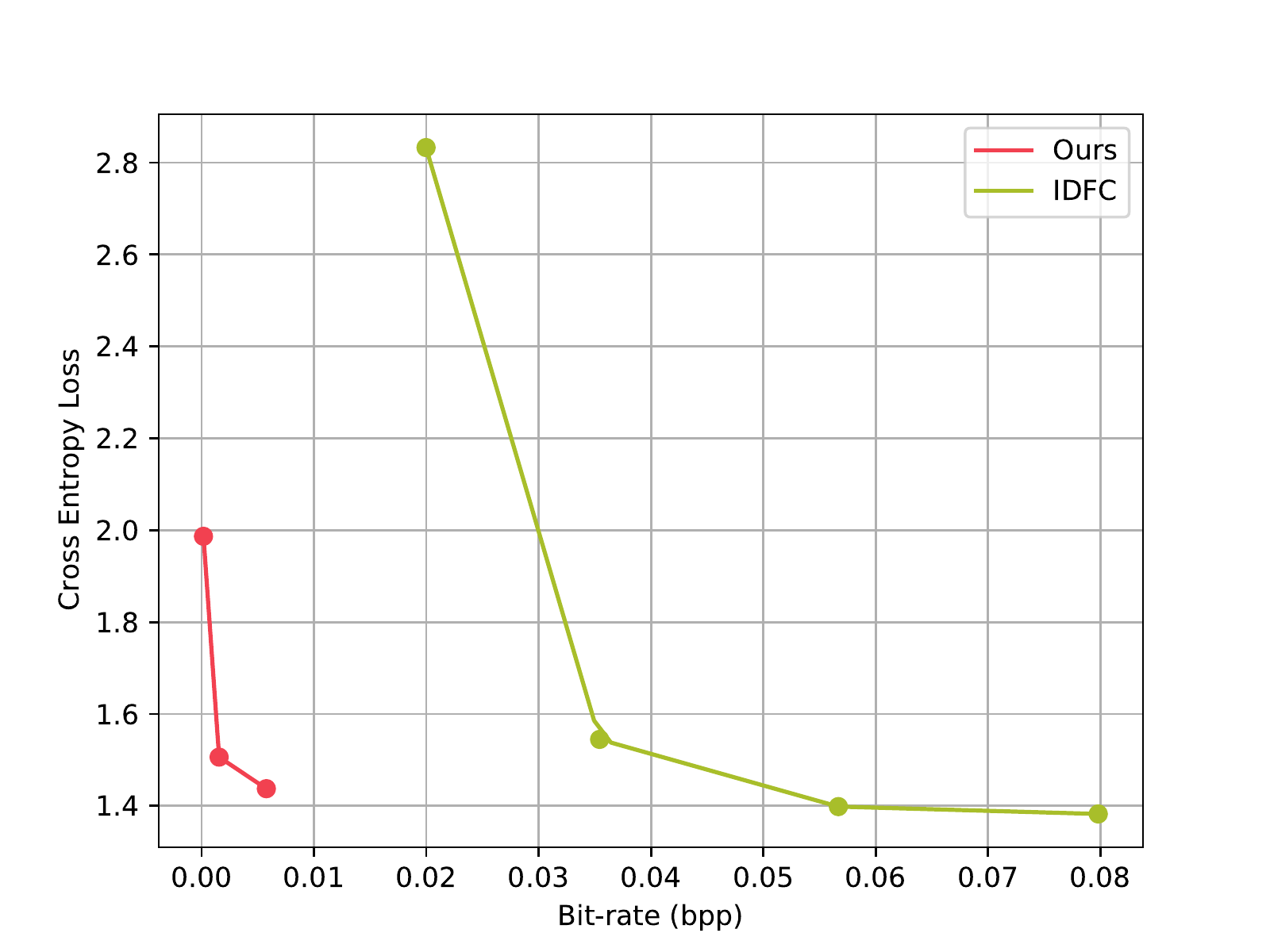}
	}\hspace{-1mm}
	\subfigure[Semantic Seg.]{
		\includegraphics[width=0.47\linewidth]{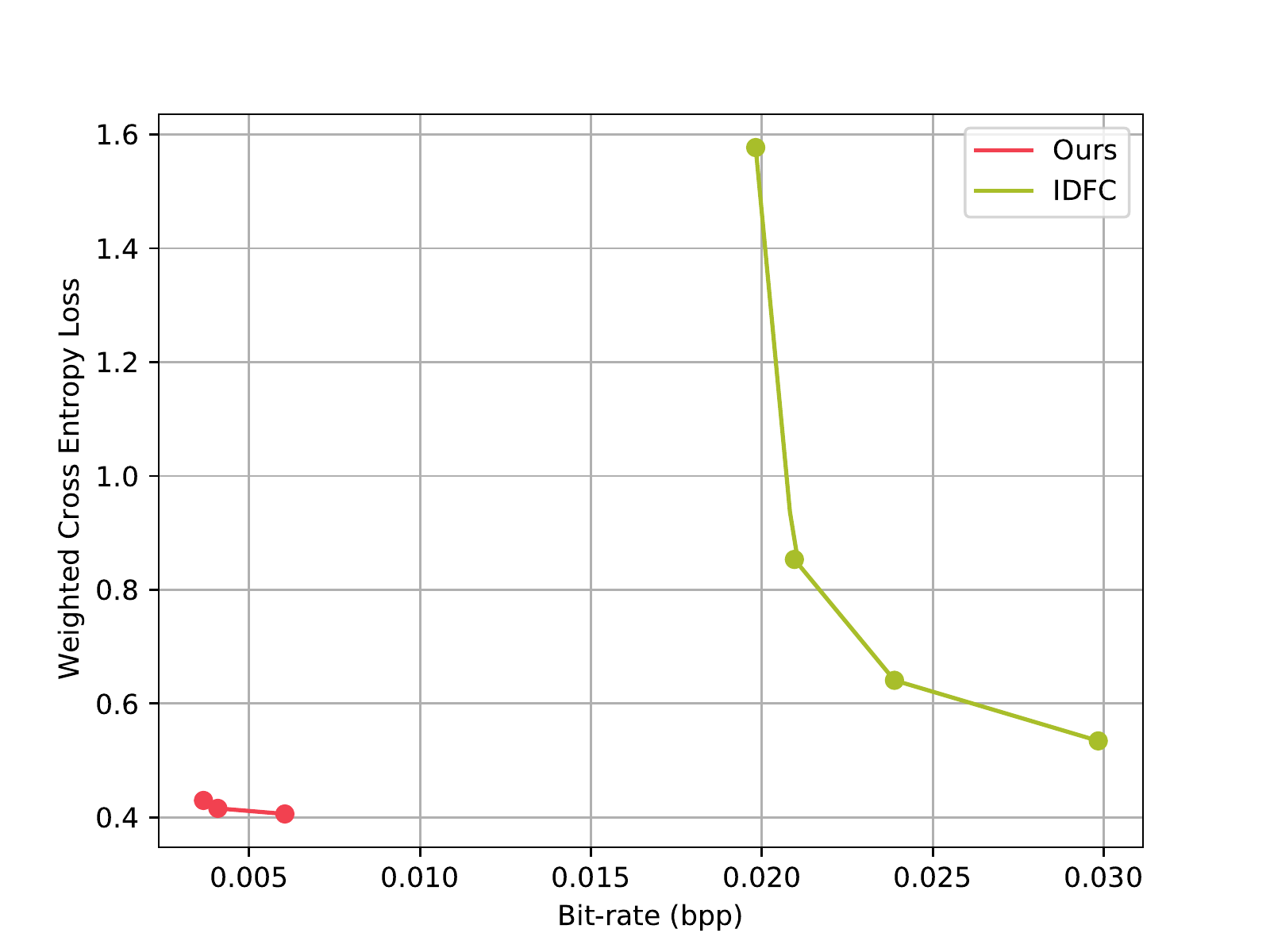}
	}
	\\ \vspace{-1mm}
	\subfigure[Surface Normal]{
		\includegraphics[width=0.47\linewidth]{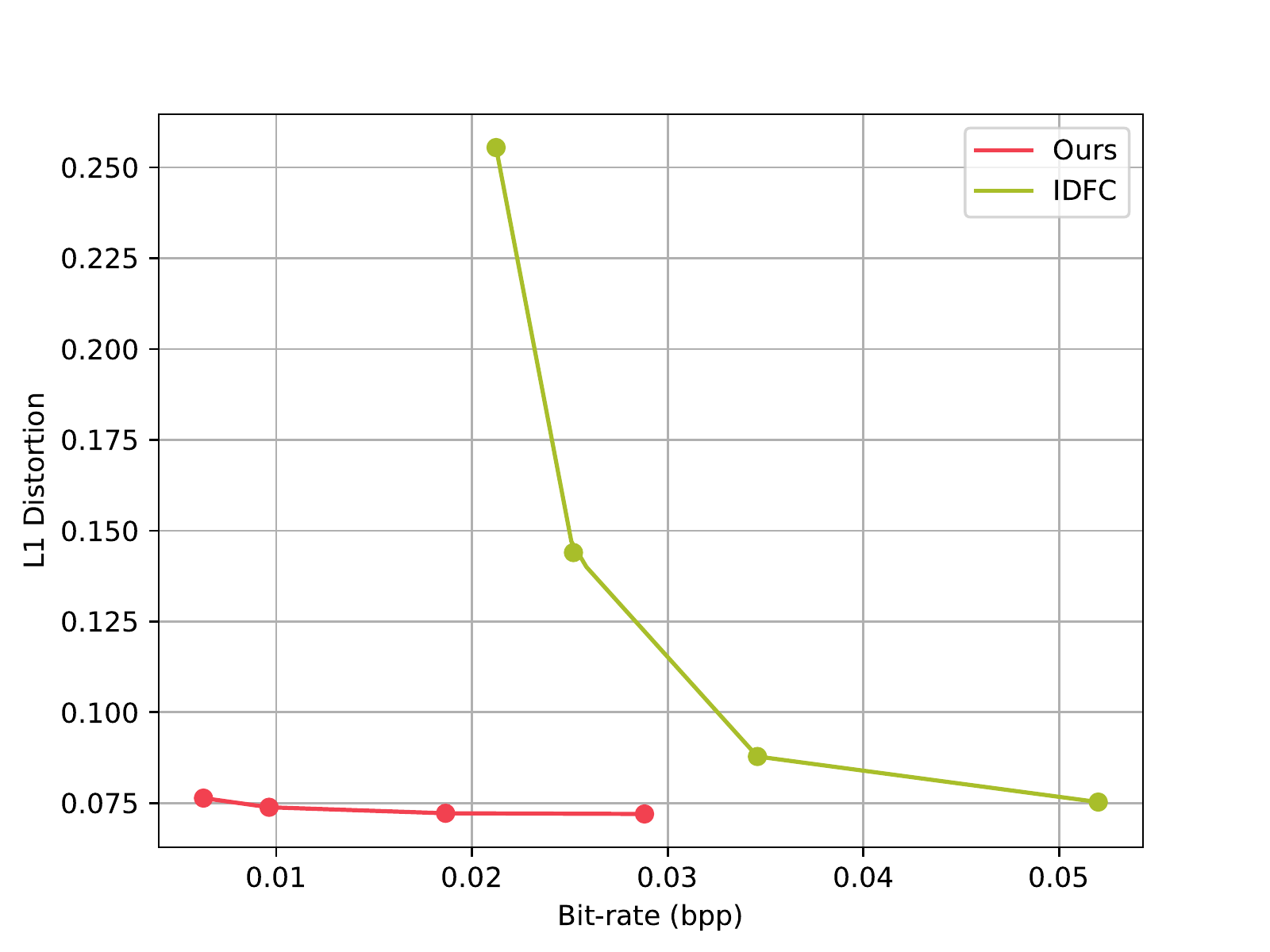}
	}\hspace{-1mm}
	\subfigure[Reshading]{
		\includegraphics[width=0.47\linewidth]{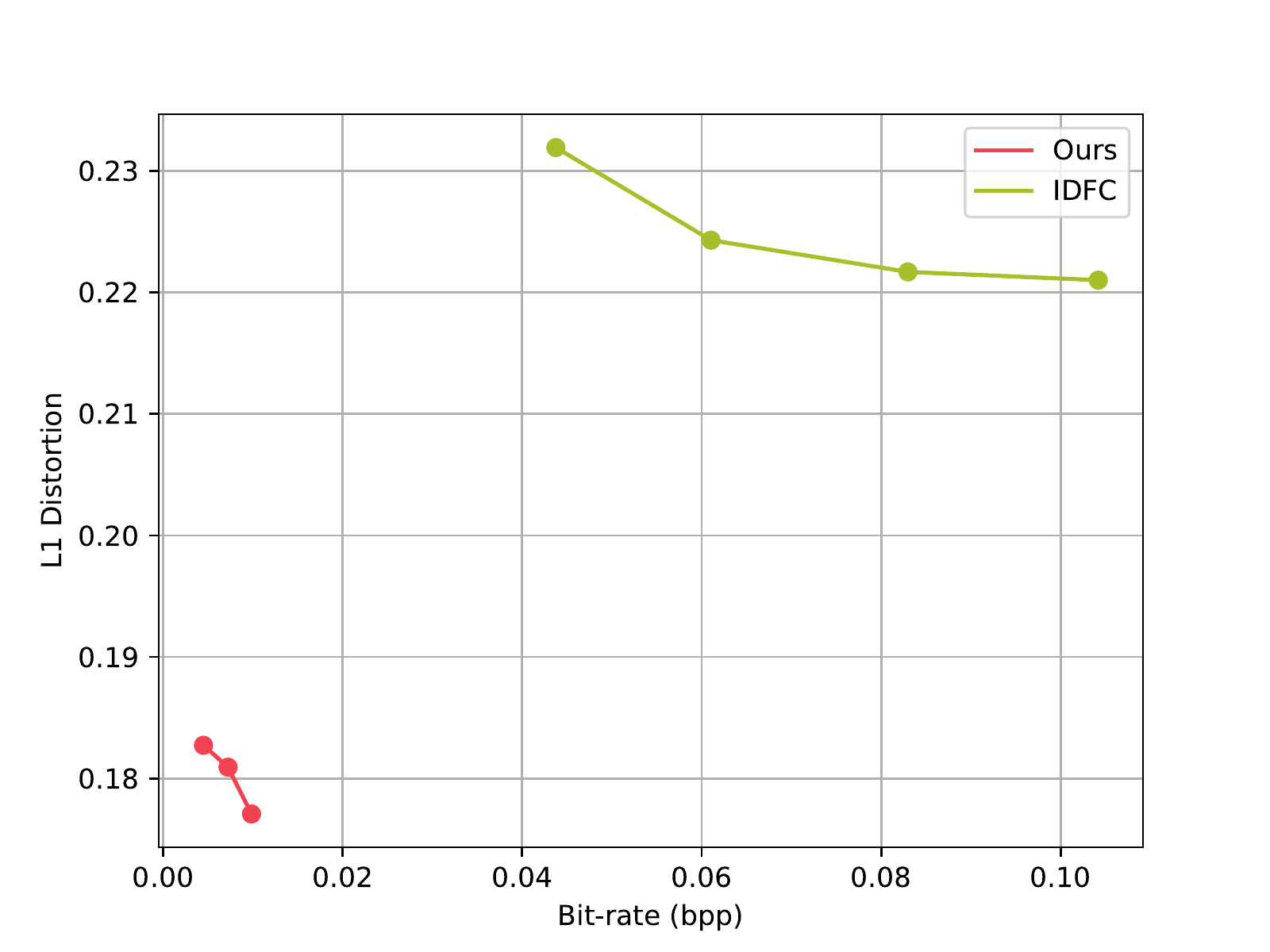}
	}
	
	\caption{Evaluation of the non-plateau bit-rate for different tasks with the proposed method and IDFC.}
	\label{fig:non-plateau}
	\end{figure}

	\begin{figure*}[htbp]
	\centering
	\subfigure[Validation $\uparrow$]{
		\includegraphics[width=0.24\linewidth]{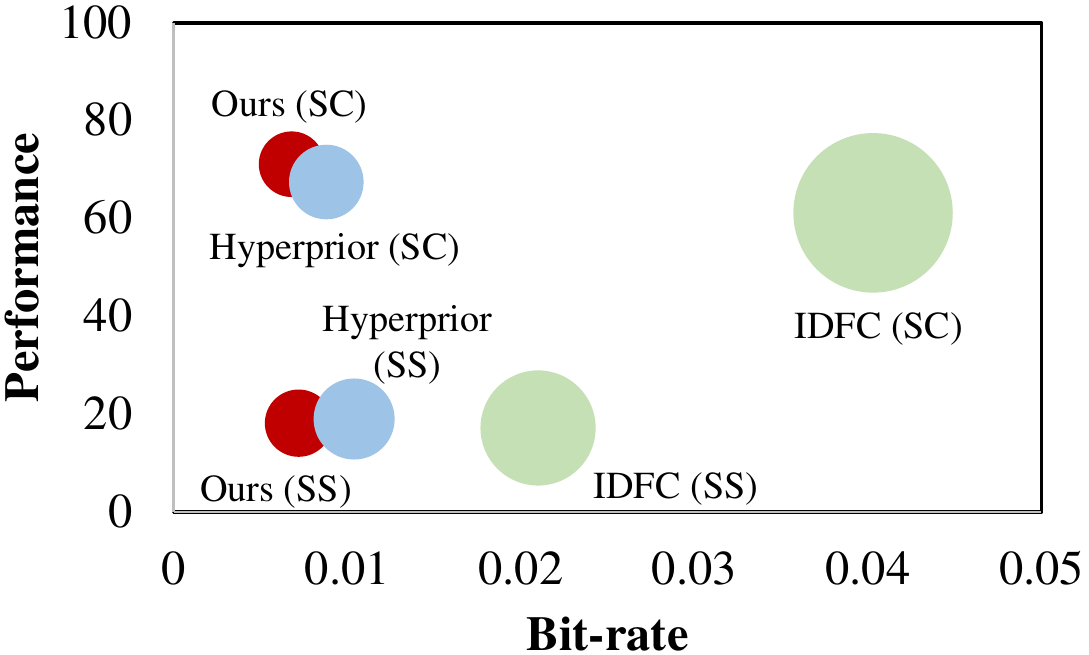}
	}
	\subfigure[Validation $\downarrow$]{
		\includegraphics[width=0.24\linewidth]{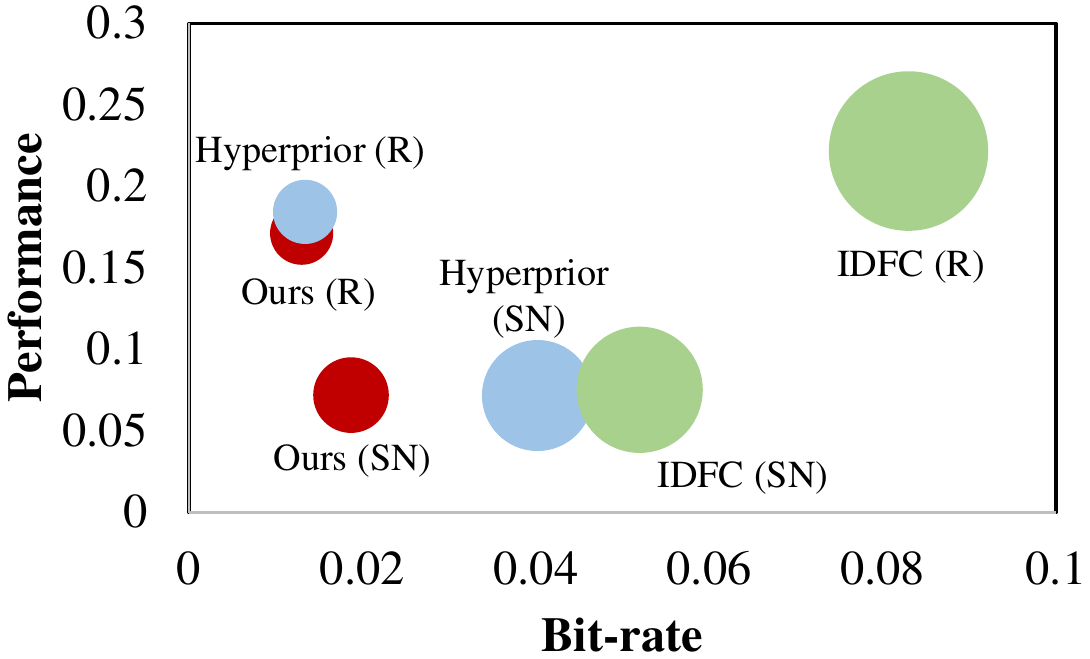}
	}
	\subfigure[Testing $\uparrow$]{
		\includegraphics[width=0.23\linewidth]{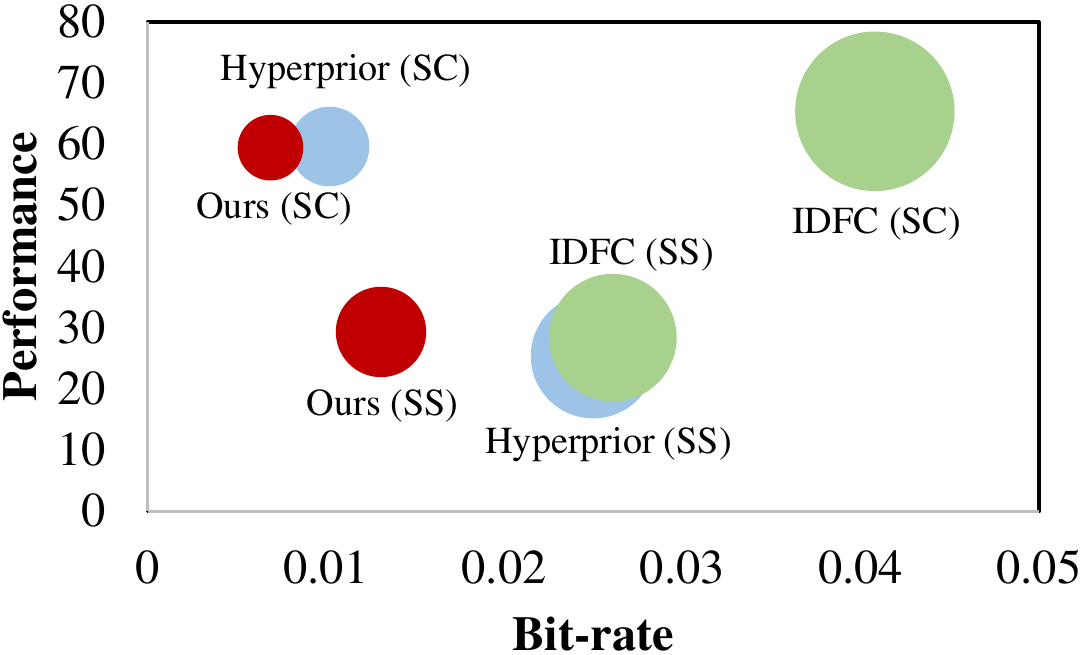}
	}
	\subfigure[Testing $\downarrow$]{
		\includegraphics[width=0.23\linewidth]{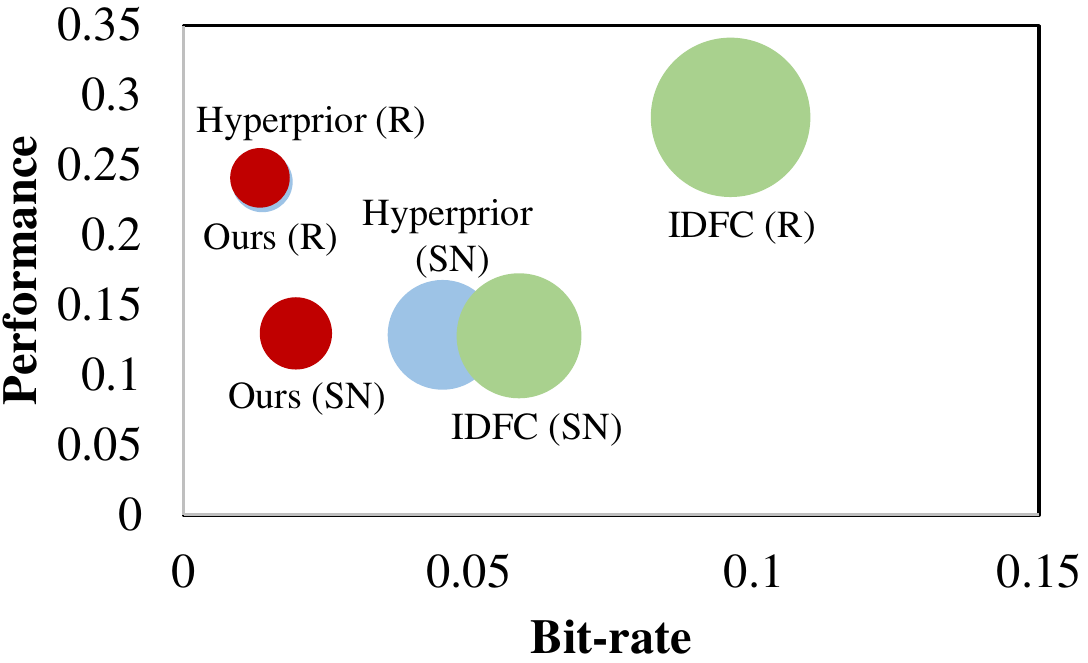}
	}
	\vspace{-4mm}
	\caption{
		The bubble charts of multiple tasks in different evaluation measures.
		SC, SS, SN, and R short for scene class, semantic segmentation, surface normal, and reshading.
		$\uparrow$ denotes that a larger number signifies a better performance while $\downarrow$ denotes that a larger number signifies a worse performance.
		\textit{The areas in bubbles visualize the bit-rate usage by different methods.}
	}
	\vspace{-3mm}
	\label{fig:radar_diff_measures}
	\end{figure*}

	\begin{table*}[htbp]
	\footnotesize
	\centering
	\caption{Analytics performance and the joint bit-rate \textit{w.r.t} different aggregation schemes. 
		The \textit{Customized}, \textit{Trinity}, \textit{Trinity*} and \textit{Hex} settings are as described in the main text.
		The best results are denoted in bold.
	}
	\begin{tabular}{cccc|cccc}
		\hline
		Task  & Metric & Original & Control Group & Customized & Trinity & {Trinity*} & Hex \\
		\hline
		Scene Class & Accuracy$\uparrow$ & 70.02\% & 75.74\% & \textbf{71.19\%} & 71.08\% & 59.64\% & 62.18\% \\
		
		Semantic Seg. & mIoU$\uparrow$  & 18.37\% & 18.85\% & 18.19\% & 18.14\% & 17.36\% & \textbf{20.30\%} \\
		
		Object Class & Accuracy$\uparrow$ & 60.17\% & 60.02\% & 61.55\% & \textbf{64.19\%} & 59.22\% & 59.75\% \\
		
		Normal & $L_1$ Distance $\downarrow$  & 0.074 & 0.071 & \textbf{0.073} & \textbf{0.073} & 0.075 & 0.074 \\
		
		Reshading & $L_1$ Distance $\downarrow$ & 0.221 & 0.172 & 0.173 & \textbf{0.168} & 0.185 & \textbf{0.168} \\
		
		Curvature & $L_1$ Distance $\downarrow$ & 0.300 & 0.296 & \textbf{0.296} & 0.299 & 0.307 & 0.306 \\
		
		Total Bit-Rate &  Bpp Sum $\downarrow$ & /     & /     & 0.059 & \textbf{0.049} & 0.050 & 0.053 \\
		\hline
	\end{tabular}%
	\label{tab:aggre}%
	\vspace{-2mm}
	\end{table*}%
	
	\begin{figure*}[t]
		\centering
		\subfigure[Trinity vs. Customized]{
			\includegraphics[width=0.31\linewidth]{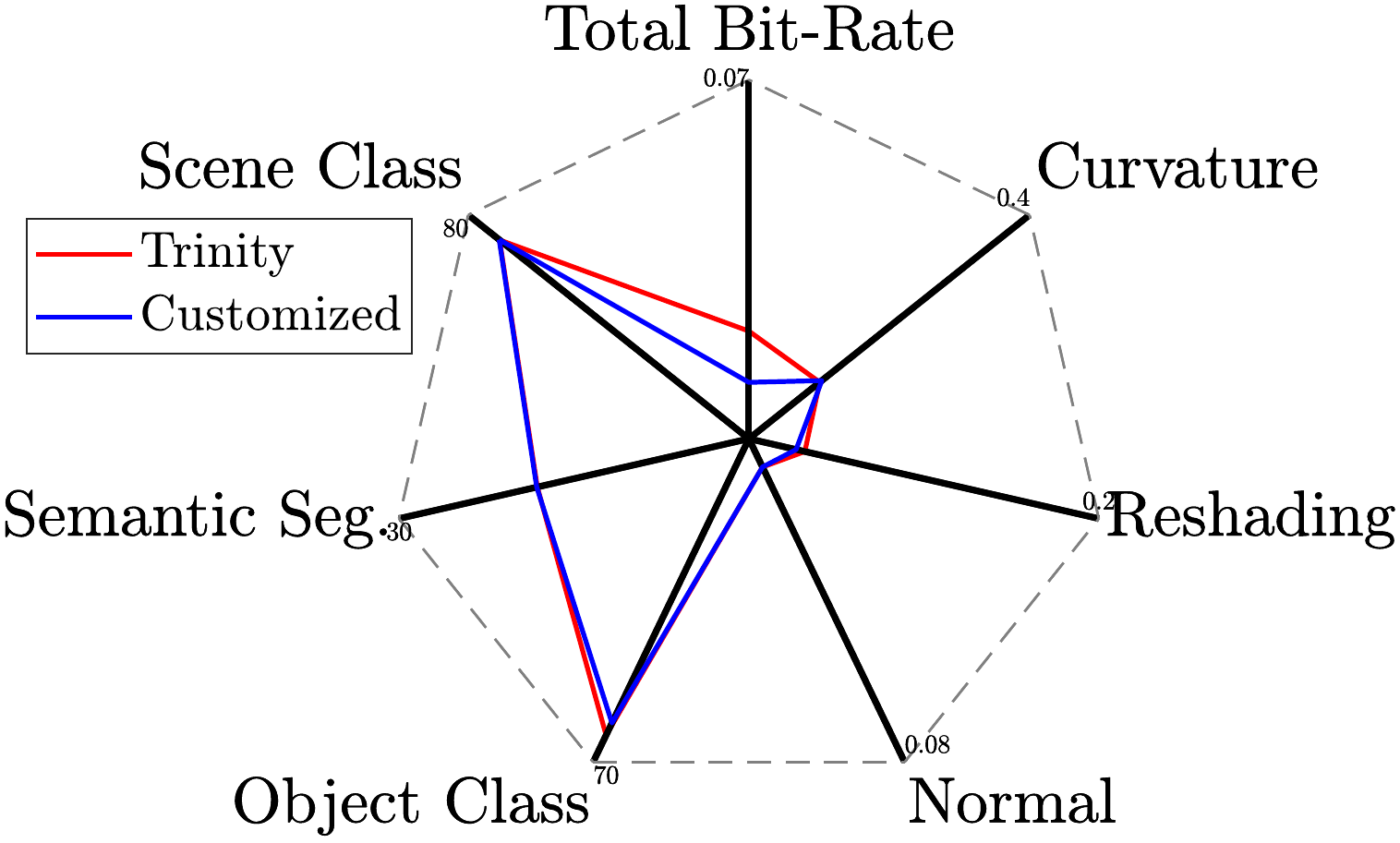}
		}
		\subfigure[Trinity vs. Trinity*]{
			\includegraphics[width=0.31\linewidth]{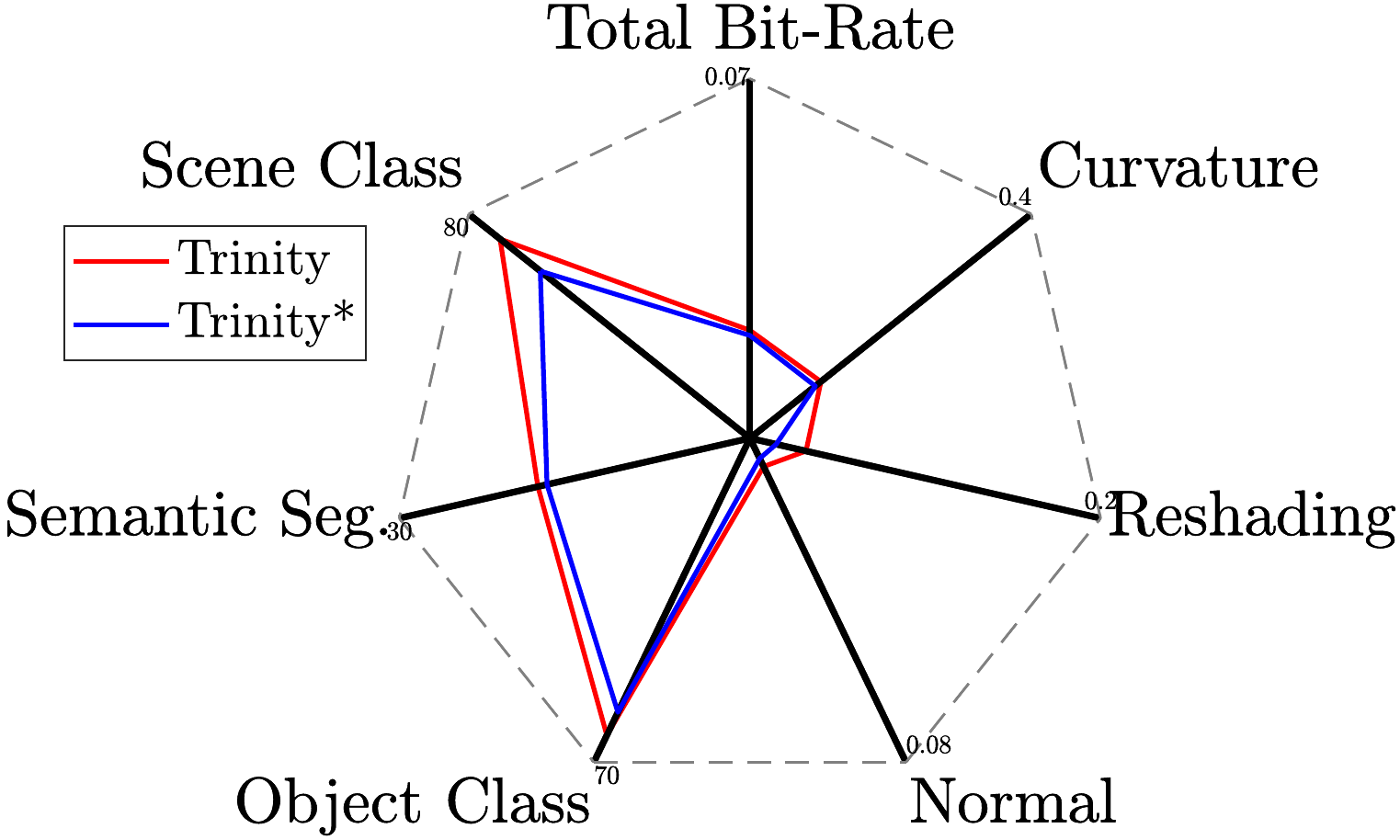}
		}
		\subfigure[Trinity vs. Hex]{
			\includegraphics[width=0.31\linewidth]{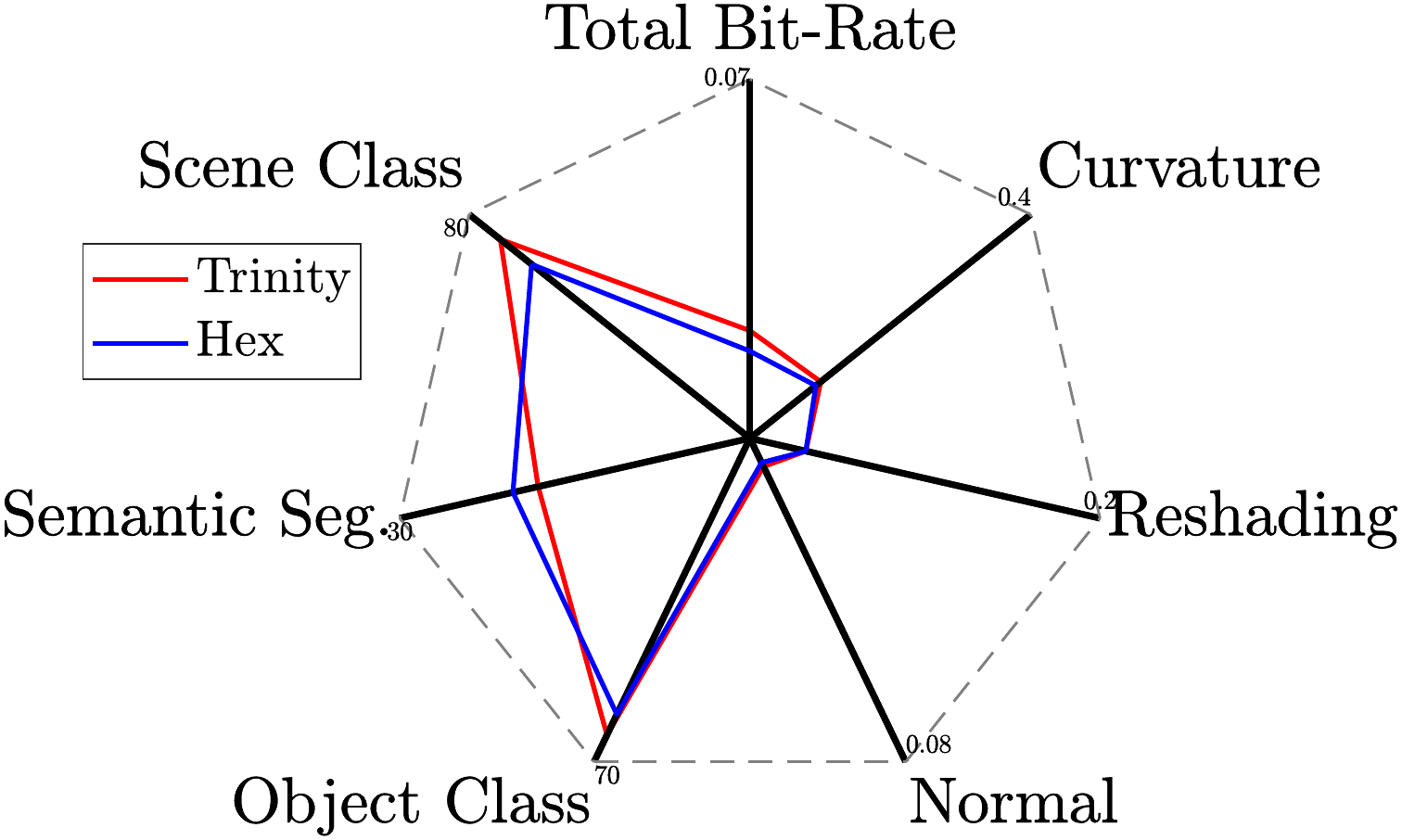}
		}
		\vspace{-3mm}
		\caption{
			The radar chart results on the multiple tasks in different evaluation measures.
			The values of the performance of normal ($x_1$), reshading ($x_2$), curvature ($x_3$), and total bit-rate ($x_4$) are adjusted via $0.08-x_1$, $0.2-x_2$, $0.4-x_3$, and $0.07-x_4$ for better visibility.
		}
		\label{fig:radar_diff_groups}
		\vspace{-4mm}
	\end{figure*}
	
	\subsection{Plateau Bit-Rate in Different Tasks}
	\label{subsec:plateau}
	
	With the proposed compression scheme, we study the plateau bit-rate \textit{w.r.t.} different tasks.
	In this experiment, we train compression models for each task, respectively, and measure the bit-rate of the compressed feature representations.
	We search for the minimal bit-rate needed to support a task to its maximally achievable performance by the provided feature, \textit{i.e.} to make the performance comparable to non-rate-control settings.
	The experiments get involved with the tasks of scene classification (Scene Class), semantic segmentation (Semantic Seg.), surface normal estimation of indoor scenes (Surface Normal), and reshading of an indoor image (Reshading).
	
	The results are shown in Table~\ref{tab:tasks}. 
	As shown, the performances of different tasks reach their plateau at different bit-rates, indicating that the information entropy to support a machine vision task varies among different tasks. 
	Image-level analytics, \textit{e.g.}, classification, requires less bit-rate to support, while pixel-level analytics require more. 
	There are also differences among pixel-level analytics.
	We also show that \textit{IDFC} consumes significantly more bit-rates. 
	Besides, as \textit{IDFC} involves a quantization based transform coding process, the quantization noise can result in unpredictable interference on the analytics performance. 
	The results suggest that such quantization noise degrades the analytics performance more significantly on the geometry related tasks. 
	Meanwhile, the proposed scheme provides better support for different kinds of tasks.
	It is noted that, as \textit{Original} is trained on the training set of the whole taskonomy dataset while other compared methods (\textit{Control Group}, \textit{Hyperprior}, and \textit{Ours}) are trained/finetuned on the training set of a subset of taskonomy dataset, the proposed method might achieve better results than \textit{Original}, which also demonstrates that our method has achieved an overall competitive performance.

	\subsection{Non-Plateau Bit-Rate in Different Tasks}
	Furthermore, we adjust the hyper-parameter $\lambda$ to explore non-plateau bit-rate with reduced performance. 
	The experiments get involved with the same four tasks as Sec.~\ref{subsec:plateau}.
	The results are shown in Fig.~\ref{fig:non-plateau}. The curves demonstrate that there exist various R-D trade-offs when coding for different machine tasks, and with the proposed compression scheme, R-D performance of different tasks outperforms \textit{IDFC} with various bit-rates.
	We also show the bubble chart results comparing \textit{IDFC}, \textit{hyperprior} and \textit{Ours} in R-D performance of various tasks in Fig.~\ref{fig:radar_diff_measures}.
	It is clearly demonstrated that, \textit{Ours} occupies the smallest areas in bubbles, which shows that \textit{Ours} generates the most compact representations.
	Besides, it is observed that, \textit{Ours} also achieves comparable or better performance than other methods.

	\subsection{(Inverse-)Transfer Mapping for Compression in Analytics Taxonomy}
	\label{sec:exp3}
	
	In this experiment, we compare the (inverse-)transfer mapping scheme with the customized group setting among multiple tasks.
	Several baselines are compared:
	\begin{itemize}
		\item \textit{Customized}: compressing feature maps for each task independently. 
		\item \textit{Hex}: jointly compressing all six kinds of representations with one model;
		\item \textit{Trinity}: an intuitively ideal compression setting that separates the six tasks into two groups for compression, \textit{i.e.} \textbf{A}: Scene Class, Semantic Seg. and Object Class; \textbf{B}: Surface Normal, Reshading and Curvature;
		\item \textit{Trinity*}: an intuitively non-optimal compression setting different from \textit{Trinity}, which separates the six tasks into two groups for compression, \textit{i.e.} \textbf{A}: Scene Class, Semantic Seg. and Object Class; \textbf{B}: Surface Normal, Reshading and Curvature.
	\end{itemize}
	The results are shown in Table~\ref{tab:aggre} and Fig.~\ref{fig:radar_diff_groups}.
	The joint compression of multiple representations saves more bit-rate.
	When all tasks reach the plateau performance, the \textit{Trinity} setting saves about 16.9\% and 19.6\% bit-rate than \textit{Customized} and \textit{Trinity*} (the last row in Table~\ref{tab:aggre}), which are also reflected in an observation from Fig.~\ref{fig:radar_diff_groups} that \textit{Trinity} owns a larger area than \textit{Customized} and \textit{Trinity*}.
	However, a larger aggregation group affects the analytics performance.
	This may be because the information from the external tasks tends to act as additional noise for the focused task.
	By grouping similar tasks in one aggregation, higher analytics performance and lower bit-rate can be achieved.
	It is noted that, \textit{Customized} does not necessarily lead to higher performance because the features of different tasks might include inter-task redundancy.
	Therefore, when the features of different tasks are optimized jointly, higher performance might be achieved by \textit{Trinity} and \textit{Hex}.
	
	
	
	
	
	\begin{table*}[t]
	\centering
	\caption{
		Analytics performance of the task relationship when the human vision need is considered.
		The best and second best results are denoted in {\color{red}red} and {\color{blue}blue}, respectively.
	}
	\vspace{-3mm}
	\begin{tabular}{ccccccc}
		\hline
		Task & Metric & Trinity & Trinity$+$ & BPG-51 & BPG-47 & BPG-43\\
		\hline
		Scene Class & Accuracy$\uparrow$
		& \color{red}{71.08\%} & \color{blue}{64.62\%} & 24.02\%& 37.14\% & 50.05\% \\
		Semantic Seg. & mIoU$\uparrow$
		& \color{red}{18.14\%} & 11.49\% & 10.08\%& 11.06\% & \color{blue}{13.93\%} \\
		Object Class & Accuracy$\uparrow$
		& \color{red}{64.19\%} & \color{blue}{60.70\%} & 33.23\%& 41.38\%& 48.04\% \\
		Reconstruction & PSNR$\uparrow$
		& / & 31.05 &29.01 & \color{blue}{31.06} & \color{red}{32.85}\\
		Bit-Rate & Bpp$\downarrow$
		& \color{red}{0.016} & 0.052 & \color{blue}{0.024} & 0.036 & 0.053\\
		\hline
		Normal & $L_1$ Distance$\downarrow$
		& \color{red}{0.073} & \color{blue}{0.081} & 0.309  & 0.231 & 0.152\\
		Reshading & $L_1$ Distance$\downarrow$
		& \color{red}{0.168} & \color{blue}{0.185} & 0.535 & 0.340 & 0.255\\
		Curvature & $L_1$ Distance$\downarrow$
		& \color{red}{0.299} & \color{blue}{0.306} & 0.331  & 0.322 & 0.313 \\
		Reconstruction & PSNR$\uparrow$
		& / & 28.79 & 29.01 & \color{blue}{31.06} & \color{red}{32.85}\\
		Bit-Rate & Bpp$\downarrow$
		&\color{blue}{0.033} & 0.054 & \color{red}{0.024} & 0.036 & 0.053\\
		\hline
	\end{tabular}
	\label{tab:humanvision}
	\vspace{-4mm}
	\end{table*}
	
	\section{Discussion}
	\label{sec:discussion}
	
	\subsection{Feature Transferability Among Tasks}
	We also explore the transferability of features among different tasks under the bit-rate constraint.
	The feature extracted by the encoder of one task is transferred, compressed and then reconstructed to the output prediction by the decoder of another task.
	The experimental results are shown in Table~\ref{tab:transfer}.
	From the results, we obtain several interesting observations:
	\begin{itemize}
		\item The best performance is achieved when the feature is encoded and decoded by the models of the same task.
		\item Surface normal is the most generalized feature and its corresponding extracted feature achieves the second best results among different tasks when being applied to handle different tasks.
		\item The features of scene class and semantic segmentation cannot be generalized to handle surface normal and reshading tasks. The bit-rates of the transferred features are lower than the other two tasks, which shows that scene class and semantic segmentation tasks include less information and cannot provide enough information to support the other two tasks.
		\item The features of surface normal and reshading can be generalized to handle scene class and semantic segmentation tasks, as they include richer information.
	\end{itemize}
	
	\begin{table}[t]
		\footnotesize
		\centering
		\caption{Evaluation of feature transferability among tasks. Each table entry include \textit{Metric} at the top and \textit{bpp} at the bottom. Adopted metrics include L$_1$ Distance for \textit{Normal} and \textit{Reshading}, Accuracy for \textit{Scene Class} and mIoU for \textit{Semantic Segmentation}.}
		\vspace{-3mm}
		\begin{tabular}{c|cccc}
			\hline
			Source Feature                  & Scene Class$\uparrow$ & \tabincell{c}{Semantic \\ Seg.$\uparrow$} & \tabincell{c}{Surface \\ Normal$\downarrow$} & Reshading$\downarrow$  \\
			\hline                                                    
			\multirow{2}{*}{Scene Class}    & 71.11\%                 & {{11.45}\%}                                & {0.1413}                                &  {0.258}          \\
			& 0.0068                  & {{0.0051}}                                 & {0.0077}                                & {0.0080}          \\
			\hline                                                    
			\multirow{2}{*}{Semantic Seg.}  & {51.27\%}            & 18.19\%                                     & {0.1470}                                 & {0.263}           \\
			& {0.0057}             & 0.0072                                      & {0.0077}                                & {0.0095}          \\
			\hline                                                    
			\multirow{2}{*}{Surface Normal} & {52.44}\%                 & 12.47\%                                     & 0.0721                                     & 0.176                \\
			& 0.0038                  & 0.0026                                      & 0.0187                                     & 0.0098               \\
			\hline                                                    
			\multirow{2}{*}{Reshading}      & 46.82\%                 & 10.42\%                                     & 0.0845                                     & 0.171                \\
			& 0.0053                  & 0.0023                                      & 0.0093                                     & 0.0130               \\
			\hline
		\end{tabular}%
		\label{tab:transfer}%
		\vspace{-3mm}
	\end{table}%
	
	
	\subsection{Task Relationship With Human Vision Task}
	We also explore feature representative capacities of different tasks when the human vision task, \textit{i.e.} full pixel reconstruction, is involved in the task aggregation.
 	The compared methods include \textit{Trinity}, \textit{Trinity$+$} (namely that the tasks in \textit{Trinity} are further grouped with the full pixel reconstruction), and \textit{BPG}, which compresses the images with BPG codec under different QP (43, 47, 51).
	The experimental results are shown in Table~\ref{tab:humanvision}.
	It is observed that, \textit{Trinity$+$} consumes a much larger bit-rate while heavily degrading the analytics performance.
	BPG can indeed provide better visual reconstruction results.
	However, the analytics performance is also harmed. 
	
	\subsection{Supporting Unseen Tasks}
	
	\begin{table}[t]
		\footnotesize
		\centering
		\caption{Evaluation of compression schemes to support unseen tasks at the plateau bit-rates.
		The best results are denoted in bold.}
		\vspace{-3mm}
		\begin{tabular}{ccc|cc}
			\hline
			Representation & bpp $\downarrow$  & Object Class $\uparrow$ & bpp $\downarrow$  & Reshading $\downarrow$ \\
			\hline
			Original & /     & 60.17\% & /     & 0.221 \\
			Internal & \textbf{0.0132} & 51.06\% & 0.0229 & \textbf{0.194} \\
			External & \textbf{0.0132} & 44.81\% & 0.0231 & 0.385 \\
			Source+ & 0.0137 & 53.50\% & \textbf{0.0167} & 0.205 \\
			BPG Image & 0.0371 & \textbf{54.56}\% & 0.0371 & 0.222  \\
			\hline
		\end{tabular}%
		\label{tab:extern}%
		\vspace{-3mm}
	\end{table}%
	
	We further explore employing the compressive representation to support external tasks that are not used in R-D training. 
	We conduct experiments with two \textit{Trinity} groups as described in Sec.~\ref{sec:exp3}, 
	while we train the compression model only for two supervision tasks.
	The representation is used to train an external decoder for an unseen task. 
	
	Following the experimental setting in Sec.~\ref{sec:exp3}, three training strategies are further adopted in comparisons:
	\begin{itemize}
		\item \textit{Internal}: When evaluating an unseen task, \textit{i.e.} object classification, the tasks \textit{in an intuitively ideal group}, \textit{i.e.} scene classification and semantic segmentation, are used for supervision.
		The same goes for the reshading task, where only the surface normal and curvature tasks are used for supervision.
		\item \textit{External}: Evaluating in supporting unseen tasks \textit{in intuitively non-optimal groups}, \textit{i.e.} supporting the reshading task (object classification) via training with the tasks of scene classification and semantic segmentation (the surface normal and curvature).
		\item \textit{Source+}: In some application scenarios, 
		although the compression component cannot be supervised by an unseen task, \textit{the pre-trained model and extracted feature at the encoder side for that task are available.}
		Thus, in this setting, the source feature for the unseen task is included in the compression but only the other two tasks are used for supervision.
		\item BPG~\cite{bpg}: the state-of-the-art image compression model that is task-independent.
	\end{itemize}
	
	The results on the validation set are shown in Table~\ref{tab:extern}.
	As shown, the proposed method can generate compressed visual representations that support external unseen tasks, 
	achieving better performance than utilizing image compression methods. 
	The results (\textit{Source+} vs. \textit{Internal} and \textit{External}) also indicate that including the additional feature representation at the encoder side can further bring in further performance gains,
	\textit{e.g.}, improving classification accuracy for object classification and reducing the bpp for reshading,
	although R-D optimization is not performed for that task.

	\section{Conclusion and Future Directions}
	\label{sec:conclusion}
	This paper formulates and summarizes the problem and solutions of video coding for machine (VCM) in recent years, targeting the collaborative optimization of compressing and transmitting multiple tasks/applications.
	Several state-of-the-art categories of methods, including features assisted coding, scalable coding, intermediate feature compression/optimization, and machine vision targeted codec, are reviewed comprehensively.
	After reviewing existing methods, we raise the new paradigm of compressive analytics taxonomy for VCM, where multi-task performance is revisited under the compression constraint.
	In particular, we propose a codebook hyperprior to compress the neural network generated features for multi-task applications/tasks.
	The codebook design helps reduce the dimensionality gap between pixels and features, and an (inverse-)transfer mapping is equipped to generate a unified compact representation.
	The experiments show the superiority of our codebook-based hyperprior model in handling multi-task applications compared to previous works, which shows a new research/solution direction for VCM.
	
	For the future researches, several pending issues are needed to pay more attention to:
	\begin{itemize}
		\item \textit{Joint optimization of video, feature and model streams}. Existing methods mainly focus on video streams and feature streams.
		As demonstrated in~\cite{lou2020tmm}, it also has the potential to involve the model in the optimization, as the knowledge might be better reused in the space of the model's parameters, which spans a more representative space and leads to discriminative capability.
		\item \textit{Theoretic investigation in the relationship of human/machine vision}. As claimed in our work, reconstruction of full pixels leads to significantly higher bit-rate usage. It is still absent on how we make a trade-off between them in different scenarios, and how they correlate and conflict with each other in theory.
		\item \textit{Consideration in Decoding Complexity}. 
		One of the most motivations in VCM is ``collaborative intelligence''~\cite{bajic2021icassp} that aims to reduce the burden at the decoder side. However, few works really embody the decoding complexity in the optimization, which shows a critical direction for the future VCM.
	\end{itemize}
	In summary, existing VCM efforts bring in abundant practices, including paradigms, solutions, techniques, and systems, and more future endeavors to improve existing methods and explore new directions are expected.
	
	{
		\footnotesize
		\bibliographystyle{ieee_fullname}
		\bibliography{egbib}
	}
	
\end{document}

%% file: summary.tex
\begin{table*}[]
	\centering
	\scriptsize
	\caption{
		An overview of video coding for machines methods in literature.
		\Checkmark~and~\XSolid\quad denote that the method owns the corresponding feature or not, respectively.
	}
	\label{tab:vcm}
	\begin{tabular}{ccccccc}
		\hline
		\multirow{2}{*}{Publication} & \multirow{2}{*}{Category} & \multirow{2}{*}{\tabincell{l}{Analytic \\ Resource}}     & \multirow{2}{*}{Output}                             & \multirow{2}{*}{Presented Task}                                        & Optimized Terms                                      &  \multirow{2}{*}{Scalable}  \\  
		& & & & & E $\|$ C $\|$ D $\|$ G $\|$ A & \\
		\hline
		Suzuki \textit{et al.} 2019~\cite{Suzuki2019icip} &  \multirow{13}{*}{\tabincell{c}{Machine \\ Vision \\ Targeted \\ Codec}}           & Image                  & Image                              & Classification                                        & $\circ$ $\|$ $\bullet$ $\|$ $\circ$ $\|$ $\circ$ $\|$ $\circ$                                                    & \XSolid       \\ \cline{0-0} \cline{3-7}
		Yang \textit{et al.} 2020~\cite{Yang2020acmmm} &             & Image                  & Image                              & Classification, Object Detection                       &$\circ$ $\|$ $\bullet$ $\|$ $\bullet$ $\|$ $\circ$ $\|$ $\circ$                                  & \XSolid       \\ \cline{0-0} \cline{3-7}
		Hou \textit{et al.} 2020~\cite{Hou2020cvpr} &             & \tabincell{c}{Color quantized \\ images}  & \tabincell{c}{Color quantized \\ images}             & Classification                                        &$\circ$ $\|$ $\bullet$ $\|$ $\bullet$ $\|$ $\circ$ $\|$ $\circ$                                  & \XSolid       \\ \cline{0-0} \cline{3-7}
		Choi \textit{et al.} 2020~\cite{Choi2020eccv} &             & Image                  & Image                              & Classification, Image Caption                         &$\circ$ $\|$ $\bullet$ $\|$ $\bullet$ $\|$ $\circ$ $\|$ $\circ$                                  & \XSolid       \\ \cline{0-0} \cline{3-7}
		Patwa \textit{et al.} 2020~\cite{Patwa2020icip} &            & Image                  & Image                              & Classification                                         & $\circ$ $\|$ $\bullet$ $\|$ $\bullet$ $\|$ $\circ$ $\|$ $\circ$                                  & \XSolid       \\ \cline{0-0} \cline{3-7}		
		Chamain \textit{et al.} 2021~\cite{Chamain2021dcc} &             & Image                  & Image                              & Detection                                             &$\circ$ $\|$ $\bullet$ $\|$ $\bullet$ $\|$ $\circ$ $\|$ $\circ$                                  & \XSolid       \\ \cline{0-0} \cline{3-7}
		Le \textit{et al.} 2021~\cite{Le2021icassp} &             & Image                  & Image                              & \tabincell{c}{ Object Detection,  \\  Instance Segmentation}             &$\circ$ $\|$ $\bullet$ $\|$ $\bullet$ $\|$ $\circ$ $\|$ $\circ$                                  & \XSolid       \\ \cline{0-0} \cline{3-7}
		Le \textit{et al.} 2021~\cite{Le2021icme} &             & Image                  & Image                              & Instance Segmentation                                 & $\circ$ $\|$ $\bullet$ $\|$ $\bullet$ $\|$ $\circ$ $\|$ $\circ$                 & \XSolid       \\
		\cline{0-0} \cline{3-7}
		Huang \textit{et al.} 2021~\cite{Huang2021icme} &             & Image                  & Image                              & \tabincell{c}{ Classification, Detection, \\  Segmentation }                & $\circ$ $\|$ $\bullet$ $\|$ $\circ$ $\|$ $\circ$ $\|$ $\circ$                                                    & \XSolid        \\		
		\hline
		Chen \textit{et al.} 2019~\cite{chen2019acmmm} & \multirow{9}{*}{\tabincell{c}{Intermediate \\ Feature \\ Compression}}  & Feature                & Feature                            & -                                                     &$\circ$ $\|$ $\bullet$ $\|$ $\bullet$ $\|$ $\circ$ $\|$ $\circ$                                  & \XSolid       
		\\ \cline{0-0} \cline{3-7}
		Chen \textit{et al.} 2020~\cite{chen2019toward} &   & Feature                & Feature                            & -                                                     &$\circ$ $\|$ $\bullet$ $\|$ $\bullet$ $\|$ $\circ$ $\|$ $\circ$                                  & \XSolid       \\ \cline{0-0} \cline{3-7}		
		Chen \textit{et al.} 2020~\cite{chen2020icip} &   & Feature                & Feature                            & -                                                     &$\circ$ $\|$ $\bullet$ $\|$ $\bullet$ $\|$ $\circ$ $\|$ $\circ$                                  & \XSolid       \\ \cline{0-0} \cline{3-7}
		Suzuki \textit{et al.} 2020~\cite{Suzuki2020icip} &   & Feature                & Feature                            & Classification                                        &$\circ$ $\|$ $\bullet$ $\|$ $\bullet$ $\|$ $\circ$ $\|$ $\circ$                                  & \XSolid       \\ \cline{0-0} \cline{3-7}
		Xing \textit{et al.} 2020~\cite{xing2020dcc} &   & Feature                & Feature                            & Action Recognition                                    &$\circ$ $\|$ $\bullet$ $\|$ $\bullet$ $\|$ $\circ$ $\|$ $\circ$                                  & \XSolid       \\ \cline{0-0} \cline{3-7}
		Choi \textit{et al.} 2020~\cite{choi2020icassp} &   & Feature                & Feature                            & Object Detection                                      &$\circ$ $\|$ $\bullet$ $\|$ $\bullet$ $\|$ $\circ$ $\|$ $\circ$                                  & \XSolid       \\ \cline{0-0} \cline{3-7}
		Hu \textit{et al.} 2020~\cite{hu2020vcip} &             & Image                  & Image                              & Classification, Image Caption                         &$\circ$ $\|$ $\bullet$ $\|$ $\bullet$ $\|$ $\circ$ $\|$ $\circ$                                  & \XSolid       \\ \cline{0-0} \cline{3-7}
		Ulhaq and Baji\'c 2021~\cite{Ulhaq2021icassp} &   & -                      & Feature                            & -                                                     & -                                                    & \XSolid       \\  \cline{0-0} \cline{3-7}
		Ikusan and Daiy 2021~\cite{Ikusan2021icme} &   & Feature                & Feature                            & Classification                                        &$\circ$ $\|$ $\bullet$ $\|$ $\bullet$ $\|$ $\circ$ $\|$ $\circ$                                  & \XSolid       \\
		\hline 
		Alvar and Baji\'c 2019~\cite{Alvar2019icip} & \multirow{9}{*}{\tabincell{c}{Intermediate \\ Feature \\ Optimization}}  & Feature                & \tabincell{c}{Semantic Map, \\ Disparity Map, Image} & \tabincell{c}{Semantic Segmentation, \\ Disparity Estimation}                    &$\circ$ $\|$ $\bullet$ $\|$ $\bullet$ $\|$ $\circ$ $\|$ $\bullet$                                  & \XSolid        \\  \cline{0-0} \cline{3-7}
		Singh \textit{et al.} 2020~\cite{Singh2020icip} &  & Feature                & Feature                            & Classification                                        &$\bullet$ $\|$ $\bullet$ $\|$ $\bullet$ $\|$ $\circ$ $\|$ $\bullet$                                  & \XSolid        \\ \cline{0-0} \cline{3-7}
		Shah and Raj 2020~\cite{Shah2020icassp} &  & Feature                & Feature                            & Classification                                        &$\bullet$ $\|$ $\bullet$ $\|$ $\bullet$ $\|$ $\circ$ $\|$ $\bullet$                                  & \XSolid        \\ \cline{0-0} \cline{3-7}
		Alvar and Baji\'c 2020~\cite{Alvar2020icassp} &  & Feature                & \tabincell{c}{Semantic Map, \\ Disparity Map, Image}                            & \tabincell{c}{Semantic Segmentation, \\ Disparity Estimation}                  &$\circ$ $\|$ $\bullet$ $\|$ $\bullet$ $\|$ $\circ$ $\|$ $\bullet$                                  & \XSolid        \\ \cline{0-0} \cline{3-7}
		Alvar and Baji\'c 2021~\cite{Alvar2021tip} &  & Feature                & \tabincell{c}{Semantic Map, \\ Disparity Map,  Image} & \tabincell{c}{ Semantic Segmentation, \\ Disparity Estimation}                    &$\circ$ $\|$ $\bullet$ $\|$ $\bullet$ $\|$ $\circ$ $\|$ $\bullet$                                  & \XSolid        \\ \cline{0-0} \cline{3-7}
		Zhang \textit{et al.} 2021~\cite{zhang2021icme} &  & Feature                & Feature                            & \tabincell{c}{Object Detection  \\ Instance Segmentation }                                      &$\bullet$ $\|$ $\bullet$ $\|$ $\bullet$ $\|$ $\circ$ $\|$ $\bullet$                                  & \XSolid        \\				
		\hline		
		Chen \textit{et al.} 2019~\cite{Chen2019icassp} &   \multirow{11}{*}{\tabincell{c}{Feature \\ Assisted \\ Coding}}  & Texture Mask                      & \tabincell{c}{Semantic Map, \\ Texture Mask, Videos} & \tabincell{c}{Texture Region}                   & $\bullet$ $\|$ $\bullet$ $\|$ $\bullet$ $\|$ $\circ$ $\|$ $\circ$                                   & \XSolid       \\ \cline{0-0} \cline{3-7}
		Li \textit{et al.} 2019~\cite{li2019adacompress} &            & -                      & Image                              & -                                                     & $\circ$ $\|$ $\bullet$ $\|$ $\bullet$ $\|$ $\circ$ $\|$ $\circ$                    & \XSolid       \\ \cline{0-0} \cline{3-7}
		Huang \textit{et al.} 2019~\cite{Chun_2019_CVPR_Workshops} &            & -                      & Image, Color Hint                  & -                                           & $\bullet$ $\|$ $\bullet$ $\|$ $\bullet$ $\|$ $\circ$ $\|$ $\circ$                                  & \XSolid       \\ \cline{0-0} \cline{3-7}
		Chang \textit{et al.} 2019~\cite{chang2019icip} &            & -                      & Edge, Image                        & -                                                     & $\bullet$ $\|$ $\bullet$ $\|$ $\bullet$ $\|$ $\circ$ $\|$ $\circ$                                  & \Checkmark      \\ \cline{0-0} \cline{3-7}
		Akbar \textit{et al.} 2019~\cite{Akbari2019icassp} &            & -           & \tabincell{c}{Semantic Map, Image, \\ Compact Image}               & -                                 & $\bullet$ $\|$ $\circ$ $\|$ $\circ$ $\|$ $\bullet$ $\|$ $\circ$                  & \Checkmark      \\ \cline{0-0} \cline{3-7}		
		Xia \textit{et al.} 2020~\cite{xia2020icme} &            & -                      & Object Mask, Image                              & -                                                          & $\bullet$ $\|$ $\bullet$ $\|$ $\bullet$ $\|$ $\circ$ $\|$ $\circ$                                  & \XSolid       \\ \cline{0-0} \cline{3-7} 
		Kim \textit{et al.} 2020~\cite{Kim2020icassp} &            & -                  & Soft Edge, Video                              & -                                                     & $\circ$ $\|$ $\circ$ $\|$ $\bullet$ $\|$ $\circ$ $\|$ $\circ$                  & \XSolid       \\ \cline{0-0} \cline{3-7}
		Prabhakar \textit{et al.} 2021~\cite{Prabhakar2021dcc} &            & -                      & \tabincell{c}{Pose, Face Mesh, \\ Video}           & -                                                     &$\bullet$ $\|$ $\bullet$ $\|$ $\bullet$ $\|$ $\circ$ $\|$ $\circ$                                  & \Checkmark      
		\\ 
		\hline
		Wang \textit{et al.} 2019~\cite{wang2019icip} &  \multirow{17}{*}{\tabincell{c}{Scalable \\ Coding}}  & Feature                & Feature, Image                     & Face Recognition                                      & $\circ$ $\|$ $\bullet$ $\|$ $\bullet$ $\|$ $\bullet$ $\|$ $\circ$ & \Checkmark      \\ 	\cline{0-0} \cline{3-7}
		
		Hu \textit{et al.} 2019~\cite{hu2020icme} &           & Image                & \tabincell{c}{Quantized Edge, \\ Color Hint, Image}                     & Facial Landmark Detection                                      & $\bullet$ $\|$ $\bullet$ $\|$ $\bullet$ $\|$ $\bullet$ $\|$ $\circ$ & \Checkmark      \\ 	\cline{0-0} \cline{3-7}
		
		Xia \textit{et al.} 2019~\cite{xia2020emerging} &           & \tabincell{c}{Sparse Points \\ and Motion}               & \tabincell{c}{Sparse Points \\ and Motion, \\ Video}                  & Action Recognition                                      & $\bullet$ $\|$ $\bullet$ $\|$ $\bullet$ $\|$ $\bullet$ $\|$ $\bullet$ & \Checkmark      \\ 	\cline{0-0} \cline{3-7}
		
		Hoang \textit{et al.} 2020~\cite{Hoang2020cvprw} &            & Semantic Map           & \tabincell{c}{Semantic Map, \\ Low-res Image, Image}      & Semantic Segmentation                                 & $\bullet$ $\|$ $\bullet$ $\|$ $\bullet$ $\|$ $\bullet$ $\|$ $\circ$ & \Checkmark      \\ \cline{0-0} \cline{3-7}	
		Yan \textit{et al.} 2020~\cite{yan2020icip} &            & Feature                & Feature                            & Classification                                        & $\bullet$ $\|$ $\bullet$ $\|$ $\bullet$ $\|$ $\bullet$ $\|$ $\bullet$ & \Checkmark      \\ \cline{0-0} \cline{3-7}
		
		Yang \textit{et al.} 2021~\cite{yang2021tmm} &           & Image                & \tabincell{c}{Quantized Edge, \\ Color Hint, Image}                     & Facial Landmark Detection                                      & $\bullet$ $\|$ $\bullet$ $\|$ $\bullet$ $\|$ $\bullet$ $\|$ $\circ$ & \Checkmark      \\ 	\cline{0-0} \cline{3-7}
		
		Wang \textit{et al.} 2021~\cite{wang2021icassp} &            & Feature                & Feature, Image                     & Face Recognition                                      & $\bullet$ $\|$ $\bullet$ $\|$ $\bullet$ $\|$ $\bullet$ $\|$ $\circ$ & \Checkmark      \\ \cline{0-0} \cline{3-7}
		Wang \textit{et al.} 2021~\cite{wang2021tmm} &            & Feature                & Feature, Image                     & Face Recognition                                      & $\bullet$ $\|$ $\bullet$ $\|$ $\bullet$ $\|$ $\bullet$ $\|$ $\circ$ & \Checkmark      \\ \cline{0-0} \cline{3-7}
		Choi and Baji\'c 2021~\cite{choi2021arxiv} &            & Feature                & Semantic Map, Image                & \tabincell{c}{Object Detection, \\ Segmentation, Reconstruction} & $\bullet$ $\|$ $\bullet$ $\|$ $\bullet$ $\|$ $\bullet$ $\|$ $\circ$ & \Checkmark      \\  \cline{0-0} \cline{3-7}
		Liu \textit{et al.} 2021~\cite{liu2021ijcv} &            & Feature                & Feature, Image                     & Classification                                        & $\circ$ $\|$ $\bullet$ $\|$ $\bullet$ $\|$ $\bullet$ $\|$ $\bullet$                  & \Checkmark      \\ \cline{0-0} \cline{3-7}
		Chang \textit{et al.} 2021~\cite{change2021icme} &    & Image                  & Semantic Map, Image                & Facial Landmark Detection                             & $\bullet$ $\|$ $\bullet$ $\|$ $\bullet$ $\|$ $\bullet$ $\|$ $\circ$ & \Checkmark      \\
		\hline
		Baji\'c \textit{et al.} 2021~\cite{bajic2021icassp} &  \multirow{2}{*}{Summary}   & --                   & --                             & --                & --                                                 & --        \\ \cline{0-0} \cline{3-7}
		Gao \textit{et al.} 2021~\cite{gao2021arxiv} &             & --                  & --                              & --                & --                                                   & --        \\ \cline{0-0} \cline{3-7}
	    \hline 
	\end{tabular}
\end{table*}